\def\reportfontpath{fonts/optimistic}
\PassOptionsToPackage{capitalize}{cleveref}
\documentclass[10pt,a4paper,authoryear]{nus-techreport}
\usepackage{xspace}

\usepackage{graphicx}
\usepackage{tikz}
\usetikzlibrary{arrows.meta,positioning,fit,calc,decorations.pathreplacing}
\usepackage{xurl}
\usepackage{array}
\usepackage{float}
\usepackage{flafter}
\usepackage{placeins}
\usepackage{multirow}
\usepackage{makecell}
\usepackage{ragged2e}
\usepackage{mathtools}
\usepackage{nicefrac}
\newif\ifteaseranimation
\teaseranimationtrue
\def\teaseranim{assets/teaser_anim}
\IfFileExists{animate.sty}{}{\teaseranimationfalse}
\IfFileExists{pdfbase.sty}{}{\teaseranimationfalse}
\IfFileExists{\teaseranim.pdf}{}{\teaseranimationfalse}
\ifteaseranimation
  \usepackage{animate}
\else
  \PackageWarningNoLine{SolarRenderer}{Teaser animation unavailable; using static poster}
\fi
\usepackage[export]{adjustbox}

\newcolumntype{P}[1]{>{\RaggedRight\arraybackslash}p{#1}}
\newcolumntype{C}[1]{>{\Centering\arraybackslash}p{#1}}
\newcolumntype{M}[1]{>{\RaggedRight\arraybackslash}m{#1}}
\newcolumntype{N}[1]{>{\Centering\arraybackslash}m{#1}}
\newcolumntype{Y}[1]{>{\hsize=#1\hsize\linewidth=\hsize\RaggedRight\arraybackslash}X}
\newcolumntype{Z}[1]{>{\hsize=#1\hsize\linewidth=\hsize\Centering\arraybackslash}X}
\setlist[itemize]{leftmargin=18pt,noitemsep,topsep=2pt}
\setlist[enumerate]{leftmargin=20pt,noitemsep,topsep=2pt}
\definecolor{originaccent}{HTML}{8B6B3F}
\colorlet{lumiaabsbg}{originaccent!10!white}
\colorlet{origintablehead}{originaccent!20!white}
\colorlet{origintablerow}{black!4}
\colorlet{oursrow}{originaccent!9!white}
\definecolor{originblue}{HTML}{245E7A}
\definecolor{originlight}{HTML}{EEF4F7}
\definecolor{verifyred}{HTML}{9C2F2F}

\newcommand{\project}{SolarRenderer\xspace}

\newcommand{\sn}[1]{\textsuperscript{\textbf{\color{originaccent}#1}}\nobreak\hspace{0.08em}}
\newtcolorbox{promptbox}[1][]{enhanced,colback=black!3,colframe=originaccent!60,
  boxrule=0.4pt,arc=2pt,left=6pt,right=6pt,top=3pt,bottom=3pt,
  before skip=6pt,after skip=6pt,fontupper=\footnotesize\ttfamily,#1}
\newtcolorbox{capcard}[1]{enhanced,colback=white,colframe=originaccent!45,
  coltitle=black,colbacktitle=origintablehead,boxrule=0.4pt,arc=2pt,
  left=4pt,right=4pt,top=2pt,bottom=3pt,toptitle=1pt,bottomtitle=1pt,
  fonttitle=\scriptsize\bfseries,fontupper=\fontsize{7.5}{8.4}\selectfont\RaggedRight,title={#1},
  before skip=0pt,after skip=0pt}

\colorlet{tabhead}{reportSecondary!9!white}
\colorlet{tabgroup}{reportSecondary!9!white}
\colorlet{tabstripe}{reportSecondary!3!white}
\colorlet{tabours}{reportAccent!9!white}
\definecolor{scorebronze}{HTML}{F7DFC9}
\definecolor{scoregold}{HTML}{F9E9B4}
\definecolor{scoresilver}{HTML}{E9EDF2}

\newcommand{\resulthead}[1]{{\fontsize{7.5}{9}\selectfont\sffamily\bfseries\shortstack{#1}}}
\newcommand{\tablefont}{\fontsize{7}{8.5}\selectfont}


\pdfmapline{+Mulish-SemiBold-t1 Mulish-SemiBold "ECEncoding ReEncodeFont" <fonts/mulish/mulish-ec.enc <fonts/mulish/Mulish-SemiBold.ttf}
\DeclareFontFamily{T1}{solarrenderermulish}{}
\DeclareFontShape{T1}{solarrenderermulish}{m}{n}{<-> fonts/mulish/Mulish-SemiBold-t1}{}
\newcommand{\findingfont}{\fontencoding{T1}\fontfamily{solarrenderermulish}\fontseries{m}\fontshape{n}\selectfont}
\definecolor{metricquality}{HTML}{AF657E}
\definecolor{metricstability}{HTML}{388F80}
\definecolor{metricstructure}{HTML}{2866A8}
\definecolor{metricfirst}{HTML}{B58A38}
\definecolor{metricjudge}{HTML}{7664A7}
\definecolor{methodrender}{HTML}{527EA6}
\definecolor{methodbase}{HTML}{647580}
\definecolor{methodaccel}{HTML}{B7854D}
\newcommand{\metriccard}[2]{%
  \par\noindent
  \tikz{\node[rounded corners=1pt,fill=#1!6,draw=none,
    inner xsep=7pt,inner ysep=3.5pt,
    text=#1!55!black,font=\findingfont\small] (metriclabel) {#2};
    \draw[#1!65,line width=1.2pt] (metriclabel.north west)--(metriclabel.south west);}%
  \par\nobreak\noindent\ignorespaces}
\newcounter{metriccardnumber}[section]
\newcommand{\numberedmetriccard}[2]{%
  \stepcounter{metriccardnumber}%
  \metriccard{#1}{\arabic{metriccardnumber}.\enspace#2}}
\newcommand{\coloredstep}[2]{%
  \par\noindent
  \textcolor{#1}{\raisebox{0.12ex}{\rule{0.45em}{0.45em}}}%
  \hspace{0.45em}\textcolor{#1!70!black}{\textbf{#2}}\ \ignorespaces}
\newtcolorbox{promptcard}[1][]{enhanced,colback=reportAccent!3,
  colframe=reportAccent!45!gray,boxrule=0.8pt,arc=8pt,
  left=11pt,right=11pt,top=8pt,bottom=8pt,before skip=0pt,after skip=0pt,
  fontupper=\ttfamily\fontsize{8.6}{10.4}\selectfont\RaggedRight,#1}
\newcommand{\promptfield}[1]{{\rmfamily\bfseries #1.}\ }

\definecolor{wtlwin}{HTML}{1BAF7A}
\definecolor{wtltie}{HTML}{CFCAC3}
\definecolor{wtlloss}{HTML}{E34948}
\newcommand{\wtlseg}[2]{\ifdim#2>0pt\textcolor{#1}{\rule[-0.15ex]{#2}{1.35ex}}\fi}

\definecolor{solarGradientStart}{HTML}{F4A02A}
\definecolor{solarGradientEnd}{HTML}{D2386C}
\reportsetup{
  accent=E8731C,resource-layout=stacked,
  title-size=25pt,title-leading=28pt,abstract-breakable=false,
  title-author-gap=10pt,author-affiliation-gap=6pt,
  author-leading=16pt,affiliation-resource-gap=8pt,resource-role-gap=6pt
}
\tcbset{nrAbstract/.append style={top=3mm,bottom=3mm,left=6mm,right=6mm,
  before skip=4pt,after skip=6pt,
  fontupper={\fontsize{10pt}{12.6pt}\selectfont}}}
\makeatletter\def\nr@abstractsize{14pt}\makeatother

\newsavebox{\solarTextBox}
\newsavebox{\solarShadedBox}

\newlength{\solarSunSize}\newlength{\solarSunGap}\newsavebox{\solarSBox}

\DeclareRobustCommand{\SolarWordmark}{%
  \texorpdfstring{\solarWordmarkBody}{SolarRenderer}}
\DeclareRobustCommand{\project}{%
  \texorpdfstring{\solarGradientText{\textbf{SolarRenderer}}}{SolarRenderer}\xspace}
\title{\texorpdfstring{\SolarWordmark: Few-Step Generative\\Reference Rendering}%
  {SolarRenderer: Few-Step Generative Reference Rendering}}
\reportauthor[1]{Guian~Fang}
\reportauthor[2]{Shengju~Qian}
\reportauthor[1]{Dongyue~Lu}
\reportauthor[3,4]{Junchao~Huang}
\reportauthor[5]{Lutao~Jiang}
\reportauthor[2]{Xin~Wang}
\reportauthor[1]{Yiren~Song}
\reportauthor[3,4]{Li~Jiang}
\reportauthor[1]{Mike~Zheng~Shou}
\makeatletter
\def\nr@authors{%
  \nr@authorformat{1}{Guian~Fang}, \nr@authorformat{2}{Shengju~Qian},
  \nr@authorformat{1}{Dongyue~Lu}, \nr@authorformat{3,4}{Junchao~Huang},
  \nr@authorformat{5}{Lutao~Jiang}\\
  \nr@authorformat{2}{Xin~Wang}, \nr@authorformat{1}{Yiren~Song},
  \nr@authorformat{3,4}{Li~Jiang}, \nr@authorformat{1}{Mike~Zheng~Shou}}
\renewcommand\nr@resourceformat[3]{{%
  \sffamily\small
  \makebox[1.2em][c]{\textcolor{reportAccent}{#1}}\hspace{0.4em}%
  \ifstrempty{#3}{#2}{\href{#3}{#2}}\par}}
\renewcommand\nr@resourceseparator{\vspace{1.5pt}}
\makeatother
\reportaffiliation[1]{NUS}
\reportaffiliation[2]{LIGHTSPEED}
\reportaffiliation[3]{CUHK(SZ)}
\reportaffiliation[4]{SLAI}
\reportaffiliation[5]{HKUST(GZ)}
\reportresource[project]{Project Page:~~\texttt{enderfga.cn/SolarRenderer}}{https://enderfga.cn/SolarRenderer}
\reportresource[code]{GitHub:~~\texttt{github.com/Enderfga/SolarRenderer}}{https://github.com/Enderfga/SolarRenderer}
\reportresource[data]{Dataset:~~\texttt{huggingface.co/datasets/Enderfga/SolarRenderer-Data}}{https://huggingface.co/datasets/Enderfga/SolarRenderer-Data}
\reportresource[models]{Weights:~~\texttt{huggingface.co/Enderfga/SolarRenderer}}{https://huggingface.co/Enderfga/SolarRenderer}
\abstract{Game engines and coding agents can specify the layout, motion, and camera
of a scene, yet turning such coarse structure into realistic video requires
a renderer that follows the structure and completes the appearance it
leaves open. Existing renderers are tied to the representation of one
engine, need tens of denoising steps or a costly distillation of their own, and learn from conditions that are
not aligned with their target videos. We present \project, a
structure-conditioned video renderer that adapts a frozen MiniMax-H3 33B
backbone with two lightweight LoRAs. Structural videos enter the native
reference pathway of the backbone at the positions of the frames they
describe, without a type identifier and described only in plain text, so that \textbf{a single control
LoRA} follows arbitrary structural representations. Starting from a
pretrained four-step LoRA, \textbf{structure-guided distribution matching}
renders in four steps, 12.3 times faster than 50-step sampling, while preserving the learned control. A
block-causal variant extends the same design to streaming and renders 768p video in \textbf{real time}. To train the model, we build a
\textbf{frame-aligned corpus} of 136K clips from game engines, simulators,
and YouTube videos. Experiments on an in-distribution test set and a
generalization benchmark of 16 structural representations show that
\project achieves state-of-the-art results across both, and generalizes
to 11 unseen representations, including pose skeletons, drawings, and point clouds.}
\AtBeginDocument{\hypersetup{%
  citecolor=reportDeepAccent,linkcolor=reportDeepAccent,urlcolor=reportDeepAccent,
  pdftitle={SolarRenderer: Few-Step Generative Reference Rendering}}}

\makeatletter

\newcommand{\solarappendix}{%
  \appendix
  \addtocontents{toc}{\protect\setcounter{tocdepth}{2}}%
  \section*{Appendix}%
  \begingroup
  \hypersetup{linkcolor=reportDeepAccent}%
  \fontsize{10}{14}\selectfont
  \renewcommand{\l@subsection}{\@dottedtocline{2}{1.5em}{2.9em}}%
  \@starttoc{toc}%
  \endgroup
  \vspace{12pt}
}
\newcommand{\solarreferences}[1]{%
  \clearpage
  \begingroup
  \small
  \bibliographystyle{solarrenderer-plainnat}%
  \bibliography{#1}%
  \endgroup
}
\makeatother

\AddToHookNext{shipout/foreground}{%
  \begin{tikzpicture}[remember picture,overlay]
    \node[anchor=south west,inner sep=0pt] at
      ([xshift=2.5cm,yshift=-2.42cm]current page.north west)
      {\includegraphics[height=1.3cm]{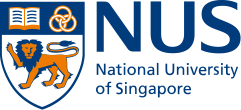}};
    \IfFileExists{assets/logo_showlab.png}{%
      \node[anchor=south east,inner sep=0pt] at
        ([xshift=-2.5cm,yshift=-2.36cm]current page.north east)
        {\includegraphics[height=1.35cm]{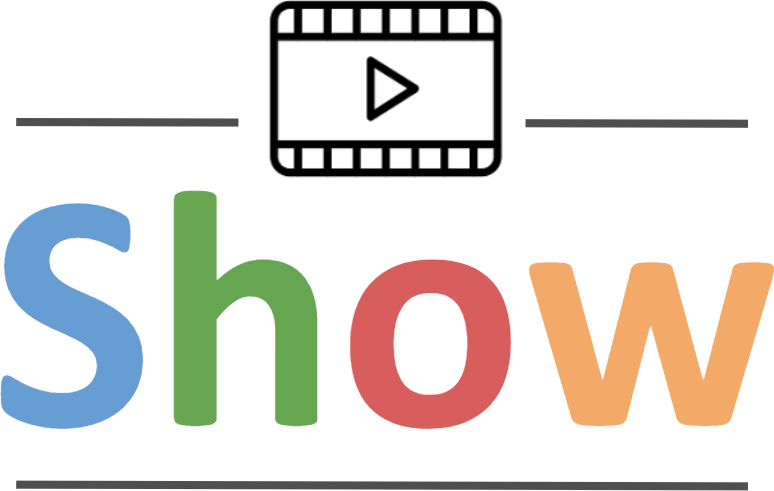}};}{}
  \end{tikzpicture}%
}

\begin{document}
  \addtocontents{toc}{\protect\setcounter{tocdepth}{-1}}%
  \enlargethispage{2.0cm}%
  \maketitle
  \vspace{-2pt}
  % Figure 1: animated teaser from one multipage PDF (\teaseranim, set in preamble.tex), every other frame at 4 fps, looping.
% Page 1 is the poster that print and most viewers show; Adobe Acrobat plays the animation on click.
\vspace{4pt}
{\centering\begin{minipage}{0.98\linewidth}\centering
  \ifteaseranimation
  \animategraphics[width=\linewidth,loop,every=2]{4}{\teaseranim}{}{}
  \else
    \includegraphics[width=\linewidth]{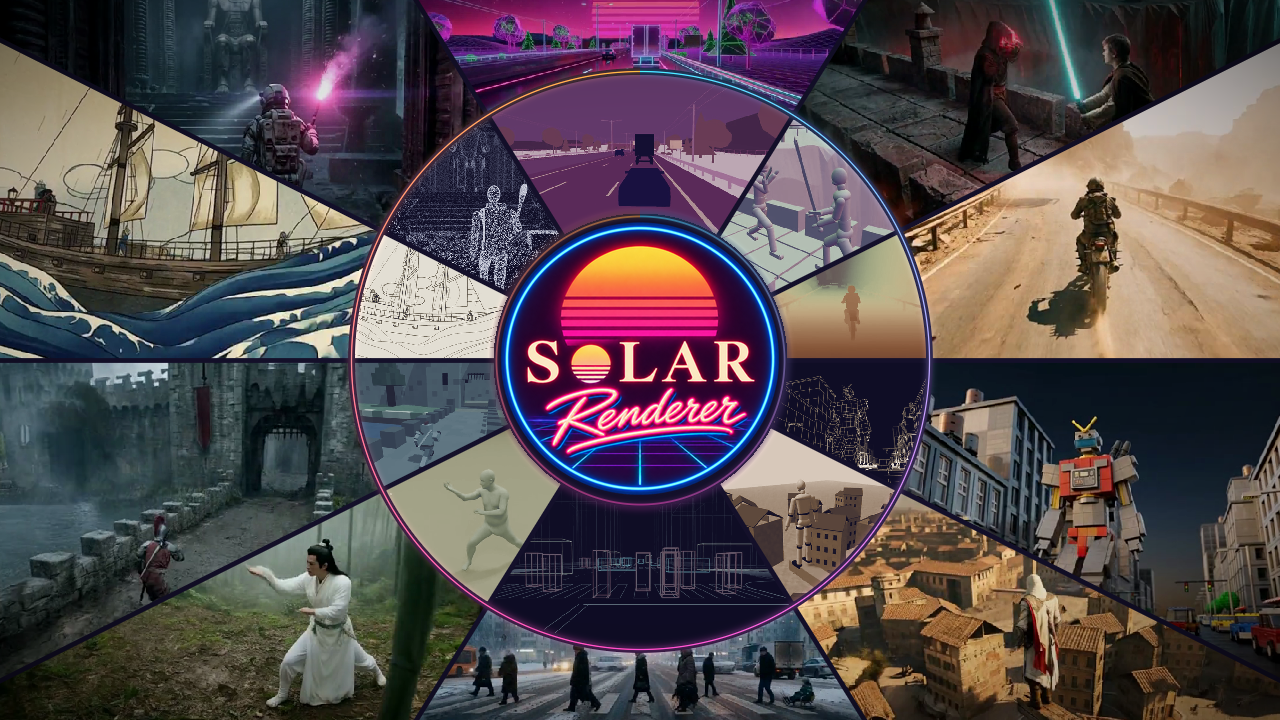}
  \fi
  \captionof{figure}{\textbf{\project renders structure of any kind into
  realistic video.} Each wedge pairs a structure (inner ring) with its
  rendering, and seven of the ten structure types are unseen in training.\ifteaseranimation\ Readers can click and play the animation
  using {\color{red}\textbf{Adobe Acrobat}}.\fi}
  \label{fig:teaser}
\end{minipage}\par}
  \reportabstract
  \clearpage

\section{Introduction}
\label{sec:introduction}

Video generation models now produce realistic footage from a text prompt
or a single image. However, visual realism alone does not ensure consistent
3D geometry, physically plausible dynamics, or temporal
coherence~\citep{huang2024vbench,bansal2025videophy2,lu2026orbia,ying2026wbench,lu2026wrbench}. In game development, film previsualization, and
agent-driven content creation, creators often specify the scene
layout, object motion, and camera trajectory in advance. Designers block out scenes, animators key character motion, and coding
agents script camera paths. These structural priors turn the task into
rendering: producing realistic video that follows the specified scene and
its evolution. With geometry and dynamics specified by these priors, the
video model can focus on appearance synthesis, reducing the scope for
structural and temporal drift. Recent systems therefore link executable world state to
video generation through proxy videos~\citep{chen2026cwm} or engine
buffers~\citep{zhan2026magpie,alaya2026renderer}, so that
\textbf{the engine determines scene evolution and a video model renders it}.

Building such a renderer still faces three problems. First, control is
\textbf{{tied to a fixed set of representations}}. Engine-driven
renderers are built around the one representation their engine exports,
such as white-box frames \citep{zhan2026magpie}, semantic maps
\citep{nvidia2026streamrender}, proxy videos \citep{chen2026cwm}, or full
G-buffers \citep{alaya2026renderer}, and general controllable generators
attach a branch per condition \citep{zhang2023controlnet,nvidia2025cosmos}
or label the type of each input
\citep{mao2026lynnrealomninativemultimodalvideo},
so a new engine or a new kind of proxy calls for a new model. Second,
rendering is \textbf{{slow}}. A large video backbone needs tens of
denoising steps, and distilling a structure-conditioned
model of this size from scratch is costly \citep{alaya2026renderer}, which
keeps such renderers far from interactive use. Third, training pairs are
\textbf{{rarely aligned frame by frame}}. Conditions estimated from
real videos carry estimator errors and differ from engine proxies
\citep{xu2026proxy2world}, and engine-paired data so far come from a single
engine.

We present \project, which adapts a frozen MiniMax-H3 33B Ref2VA backbone
\citep{minimax2026h3} and addresses the three problems in turn (\cref{fig:pipeline}). For \textbf{{omni-structure control}},
structural videos enter the native reference pathway of the backbone, each structural latent takes the RoPE \citep{su2021roformer} position of
the target frame it describes, and no reference carries a type identifier,
as each representation is described in plain text instead,
so that a single control LoRA \citep{hu2021lora} follows block-outs, semantic maps, shaded
geometry, and body meshes alike. For \textbf{{fast rendering}}, we post-train a pretrained
four-step LoRA of the same backbone with structure-guided distribution matching, which extends distribution
matching distillation \citep{yin2024dmd} with a teacher extrapolated along the structural condition, so that the model renders in four steps, 12.3 times faster than 50-step sampling
while preserving the learned control. A block-causal variant of the same
design further streams 768p video in \textbf{real time}. For
\textbf{{frame-aligned supervision}}, we collect 136K clips from
eight sources whose structure is exported by game engines and simulators or
fitted to YouTube videos, so that each structural frame matches
its target frame in both content and timing.

We evaluate \project on an in-distribution test set with ground-truth
videos and on a generalization benchmark that covers 16 structural
representations, 11 of them unseen in training. The results show that our four-step model outperforms existing acceleration LoRAs and the official
backbone, surpasses existing structure-conditioned renderers, and achieves
a new state of the art in structure-conditioned video
rendering. Notably, our method shows \textbf{\textit{strong generalization to
unseen structure}}. It faithfully renders pose skeletons, pencil
storyboards, children's crayon drawings, wireframes, and point clouds,
none of which appears in training. Our contributions are fourfold:
\begin{itemize}[itemsep=3pt]
  \item We present a renderer that handles \textbf{any structural
  representation} with a single control LoRA through the native reference
  pathway, without type identifiers or added branches.
  \item We propose \textbf{structure-guided distribution matching}, which
  post-trains a pretrained four-step LoRA so that our model renders in four steps while preserving its structural control.
  \item We build a \textbf{frame-aligned paired corpus} of 136K clips from
  game engines, simulators, and YouTube videos, in which every structural
  frame is aligned with its target frame.
  \item Our method achieves \textbf{state-of-the-art results} on an
  in-distribution test set and a generalization benchmark of 16 structural
  representations, including 11 unseen in training, and its causal variant
  renders 768p video in \textbf{real time}.
\end{itemize}

\section{Related Work}
\label{sec:related-work}

\noindent\textbf{Rendering Engine and Agent State into Video.}
World modeling encompasses 3D and 4D scene generation~\citep{kong2025worldmodelingsurvey}
and interactive proxies that provide feedback for agent
learning~\citep{yang2026quovadis}.
Recent systems let a game engine, a simulator, or a coding agent decide what
happens in a scene and let a video model render it. Most of them train a
renderer on one structural representation exported by their engine, such as
white-box frames~\citep{zhan2026magpie,gomeznogales2026coarsetoreal}, semantic
maps~\citep{nvidia2026streamrender}, depth and normal
maps~\citep{chi2026agentgarten}, projected 3D
boxes~\citep{huang2026programmable}, or the proxy video of a code world
model~\citep{chen2026cwm,liao2026streamingcwm}, and several distill the
renderer to a few streaming steps for real-time
use~\citep{vast2026edenlive,nvidia2026streamrender}. A
related branch renders from full G-buffers that also carry materials and
lighting~\citep{zeng2026rgbx,alaya2026renderer,huang2026worldrenderer,liang2025diffusionrenderer}.
Other works avoid paired proxy data, either by learning from posed RGB-D
video~\citep{xu2026proxy2world} or by showing proxy renders to a frozen model
during the early denoising steps only~\citep{liang2026codevideomodel}.
However, each renderer is built around the representation of its own
engine, so a new engine or a new kind of proxy calls for a new renderer.
\project follows many kinds of structure with \textbf{\textit{a single
control LoRA}},
including structure it has never seen.

\noindent\textbf{Structure-Conditioned Video Generation.}
Controllable video generation commonly injects conditions in one of two
ways. Branch-based methods attach a trainable copy or added control blocks
to the backbone, from ControlNet~\citep{zhang2023controlnet} to
multi-condition variants on video
backbones~\citep{pai2026controlnetunion,nvidia2025cosmos,nvidia2025worldsimulation,jiang2025vace}.
Sequence-based methods place condition tokens next to the target tokens and
read them with full attention, often through a small
LoRA~\citep{tan2024ominicontrol,ju2025fulldit,fang2025framepromptincontextcontrollableanimation,mao2026lynnrealomninativemultimodalvideo,tang2026paiactor,gao2026paistudio}.
Further works steer generation through warped noise, 3D tracking videos,
reference videos, predicted LoRAs, or jointly generated intrinsic
channels~\citep{burgert2025gowiththeflow,gu2025das,bian2026videoasprompt,wu2026video2lora,xi2025ctrlvdiff,lu2026see4d},
and training-free methods condition an existing model through structured
first frames or a ControlNet-style world
model~\citep{li2025anyi2v,wang2025zeroshotrealism}. However, these methods
add parameters for control, label the type of each condition, or are
trained on a fixed set of condition kinds. \project passes an \textbf{\textit{untyped structural
video through the native reference pathway}} of the backbone with one LoRA,
so that the same model also follows structure unseen in training.

\noindent\textbf{Few-Step Distillation and Acceleration Adapters.}
Few-step video generation is usually obtained in two ways. One distills a
multi-step teacher into a few-step student by distribution
matching~\citep{yin2024dmd,yin2024dmd2,liu2026decoupleddmd,wang2026pdmd,yu2026dmad,huang2025selfforcing,zhao2026maskforcing}
or consistency and flow-map distillation~\citep{zheng2026rcm,gu2026anyflow}. The other
trains a plug-in acceleration LoRA once on a base model and attaches it to
models derived from the same
base~\citep{luo2023lcmlora,lin2024animatedifflightning,longlive2026plug,fastvideo2026fasth3},
with recent work making such LoRAs transferable between base and distilled
models~\citep{li2026dart,wang2026casa}. For conditional models, several
works distill the conditional model itself or re-tune its control module
after distillation~\citep{wang2026liveedit,li2026flashmotion,sun2026streamavatar},
and MotionEcho~\citep{rong2025motionecho} restores motion control at
inference with guidance from the slow teacher. However, distilling a large
control model from scratch is costly, and a generic few-step adapter attached
unchanged lowers the visual quality and first-frame fidelity of the control
model. \project \textbf{\textit{post-trains a four-step LoRA}} on
the frozen control model, with a distribution-matching teacher extrapolated
toward its structure-conditioned prediction.

\section{\project: Native Reference Rendering}
\label{sec:method}

\project turns a pretrained video backbone into a renderer by adding only
lightweight adapters. In this section, we formulate the rendering task and
outline the design (\cref{sec:task}), describe how structural references
are aligned with the target and described in text
(\cref{sec:conditioning}), and present the two-stage training of control
and acceleration (\cref{sec:training}), followed by its causal streaming
extension (\cref{sec:causal}).

\begin{figure}[t]
  \centering
  \includegraphics[width=\linewidth]{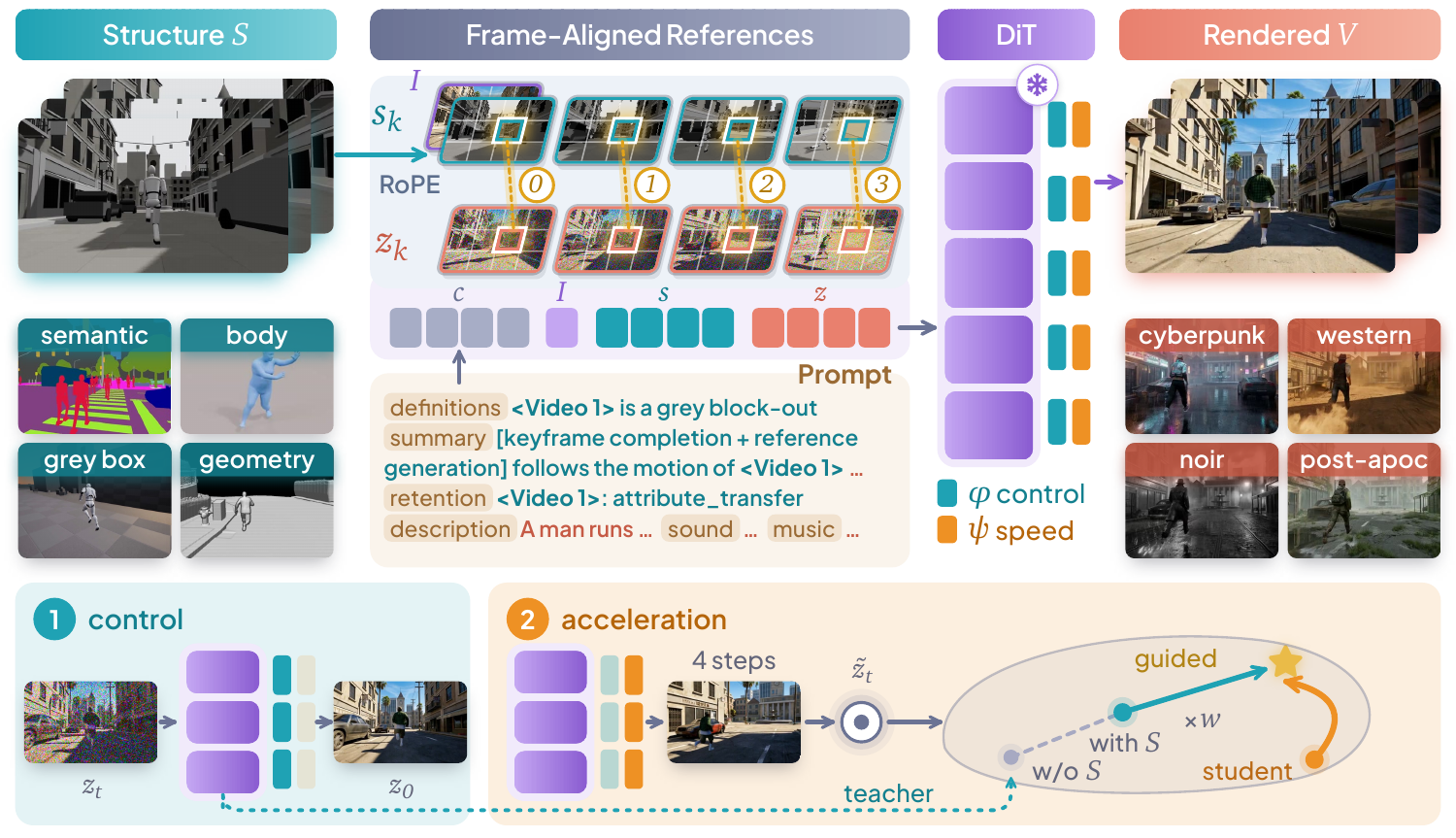}
  \caption{\textbf{\project reads any structural video through the native reference pathway of a frozen DiT.} The sequence of the single-stream DiT holds text, the optional first frame $I$,
  structural latents $s_k$, and noisy latents $z_k$, each $s_k$ taking the
  RoPE position $(t,h,w)$ of $z_k$ and $I$ index~0.
  \sn{1}~The control LoRA $\phi$ is trained by flow matching.
  \sn{2}~The acceleration LoRA $\psi$ is distilled
  toward the teacher with $S$, extrapolated by $w$ from the one
  without $S$.}
  \label{fig:pipeline}
\end{figure}

\subsection{Problem Formulation and Overview}
\label{sec:task}

\noindent\textbf{Structure-Conditioned Rendering.}
Let $V=(v_1,\ldots,v_T)$ be a target video and $S=(s_1,\ldots,s_T)$ a
structural video rendered from the same camera at the same timestamps.
Depending on the source, $S$ is a grey block-out, a semantic map, a depth
map, object masks with skeletons, or a body mesh. Given $S$, an optional
first frame $I$, and a text description $c$, \project models
\begin{equation}
  p_{\theta}(V\mid S,I,c),
  \label{eq:task}
\end{equation}
where $S$ or $I$ may be absent while $c$ is always provided. This yields
four input combinations, namely \sn{1}\textit{structure with a first
frame}, \sn{2}\textit{structure alone}, \sn{3}\textit{a first frame
alone}, and \sn{4}\textit{text alone}. The structural video specifies
\textbf{\textit{the scene state}}, including the layout, geometry,
character bodies, motion, and camera of every frame, but leaves materials,
illumination, and tiny structures undetermined. Rendering is therefore a
\textbf{\textit{constrained completion}} problem, in which the output
must agree with $S$ wherever $S$ is informative, while $I$, $c$, and the
pretrained prior supply the remaining appearance.

\noindent\textbf{Design Overview.}
A large video backbone already provides a strong appearance prior, so
\project keeps the backbone frozen and adds only lightweight adapters
(\cref{fig:pipeline}). Structural references enter through the native
reference pathway with target-aligned positions, and fixed sentences in
the prompt describe each reference. A control LoRA follows these references, and an acceleration LoRA from the same backbone enables few-step rendering.

\subsection{Frame-Aligned Reference Conditioning}
\label{sec:conditioning}

\noindent\textbf{Native Reference Pathway.}
A renderer must match every structural frame with the target frame it
describes. Instead of attaching a ControlNet branch
\citep{zhang2023controlnet} or concatenating channels, both of which
introduce parameters trained from scratch, we feed the structural video
and the optional first frame through the reference pathway of the
backbone \citep{minimax2026h3}. This pathway packs text, reference videos,
audio, and the target video into one full-attention sequence
(\cref{fig:pipeline}), so the control adapter can
\textbf{\textit{inherit the pretrained ability to read reference
videos}}. We refer to the tokens of the supplied references as
\emph{reference tokens} and to those of the generated video as
\emph{target tokens}. Yet native reference tokens occupy a separate
temporal segment before the clip and their own spatial grid, so the
zero-shot backbone uses a structural video as a loose reference and
copies it or changes the camera instead of following
the structure.

\noindent\textbf{Frame-Aligned Positional Encoding.}
Since rotary position embedding \citep{su2021roformer} makes attention
depend on the relative offset between tokens, we supply the missing
correspondence through positions. Each reference token receives
\textbf{\textit{the RoPE coordinates of the target token it describes}}.
The $k$-th structural latent frame shares the temporal index and spatial
grid of the $k$-th target latent frame, and the first frame is encoded as
a static clip at index~0 (\cref{fig:pipeline}). Tokens at the same position remain distinguishable through their
timesteps, as reference tokens keep the near-clean value that the backbone
assigns to reference inputs while target tokens carry the current noise
level. This extends shared positions for aligned image conditions
\citep{tan2024ominicontrol} to temporally aligned video and, like
in-context control of video models without structural changes
\citep{fang2025framepromptincontextcontrollableanimation}, requires no
additional module.

\noindent\textbf{Untyped References with Fixed Descriptions.}
\label{sec:prompts}
Structural representations from different sources constrain different
aspects of the target (\cref{sec:sources}). We keep reference tokens
\textbf{\textit{untyped}} in the sequence, without a type identifier or a
modality embedding, so the adapter infers what a reference constrains from
the structural video itself. The prompt additionally describes each
representation with two fixed sentences, a definition of what it shows and
a rule that names the attributes to follow and the visual properties that
must not appear. For a scene block-out:
\begin{promptbox}
<Video 1> is an untextured grey block-out of the scene, rendered from the
same camera as the target video, giving the layout, the large geometry, the
characters' bodies, their motion and the camera of every frame.\\[2pt]
<Video 1> (frame-by-frame layout, geometry, motion and camera):
attribute\_transfer - the layout, large shapes, character bodies, motion
and camera are followed exactly; small objects, materials, colours and
lighting are not given by <Video 1> and are the target video's own; the
grey look of <Video 1> is not shown.
\end{promptbox}
The same representation always receives the same sentences. Open-vocabulary text
describes each representation, so a new format needs only a new
description, with no added parameters or retraining.

\subsection{Decoupled Two-Stage Training}
\label{sec:training}

\project should render with few sampling steps while retaining the
control learned at full steps. Training both objectives together does not
achieve this, because whenever the acceleration objective can update the
control adapter, it also alters how the structure is followed. Distilling
a few-step student from scratch is costly at this scale and prone to
instability \citep{alaya2026renderer}. We therefore
\textbf{\textit{decouple control and acceleration}} into two stages on
one frozen backbone, which also preserves its visual prior and keeps it
compatible with adapters trained on the same weights.

\noindent\textbf{Stage 1: Control Adaptation.}
We train a control LoRA \citep{hu2021lora} with the flow-matching
objective. Each adapted linear layer adds a low-rank update
to its frozen pretrained weight $W_0$,
\begin{equation}
 W = W_0 + \Delta W_{\mathrm{ctrl}},\qquad
 \Delta W_{\mathrm{ctrl}}=\frac{\alpha}{r}BA,
 \label{eq:lora}
\end{equation}
where only the factors $B$ and $A$ of rank $r$ are trained and $\alpha/r$
scales the update. The adapter is applied to the attention and
feed-forward projections of every block. Since the backbone is a
\textbf{\textit{single-stream transformer}}, the same adapted projections
process reference and target tokens, so one adapter learns both to
interpret the structure and to render from it. With a clean target latent
$z_0$, Gaussian noise $\epsilon$, and noise coordinate $t$, the objective
\citep{lipman2022flow} is
\begin{equation}
 z_t=(1-t)z_0+t\epsilon,\qquad
 \mathcal{L}_{\mathrm{FM}}
 =\mathbb{E}_{z_0,\epsilon,t,c,S,I}
 \left\|f_{\theta_0,\phi}\bigl(z_t,t,c,R(S,I)\bigr)-(\epsilon-z_0)\right\|_2^2,
 \label{eq:fm}
\end{equation}
where $\theta_0$ denotes the frozen backbone, $\phi$ the control adapter,
and $R(S,I)$ the available references. The loss is computed on the target
video tokens only.

\noindent\textbf{Stage 2: Acceleration Adaptation.}
\label{sec:few-step}
Because the backbone stays frozen, a pretrained four-step LoRA of the same
backbone composes directly with the control adapter. We start from the
rank-128 four-step LoRA of DMAD \citep{yu2026dmad}, which predicts a clean
sample at every step and re-noises it for the next. It initializes the first
128 ranks of a rank-384 LoRA whose remaining ranks start at zero update, so
training starts from DMAD's update with room to grow, and we compose it as
\begin{equation}
 W = W_0+\Delta W_{\mathrm{ctrl}}+\Delta W_{\mathrm{acc}}.
 \label{eq:composition}
\end{equation}
The composition already follows the structure, but the acceleration
adapter was distilled for the backbone alone and has never seen the
control model. The control adapter $\phi$ determines what is rendered from
the structure, whereas the acceleration adapter $\psi$ determines how four
denoising steps reach the result. We therefore update only $\psi$ from its
pretrained weights and keep $\phi$ frozen, so that \textbf{\textit{the
control is preserved}}.

\begin{figure*}[t]
  \centering
  \includegraphics[width=\linewidth]{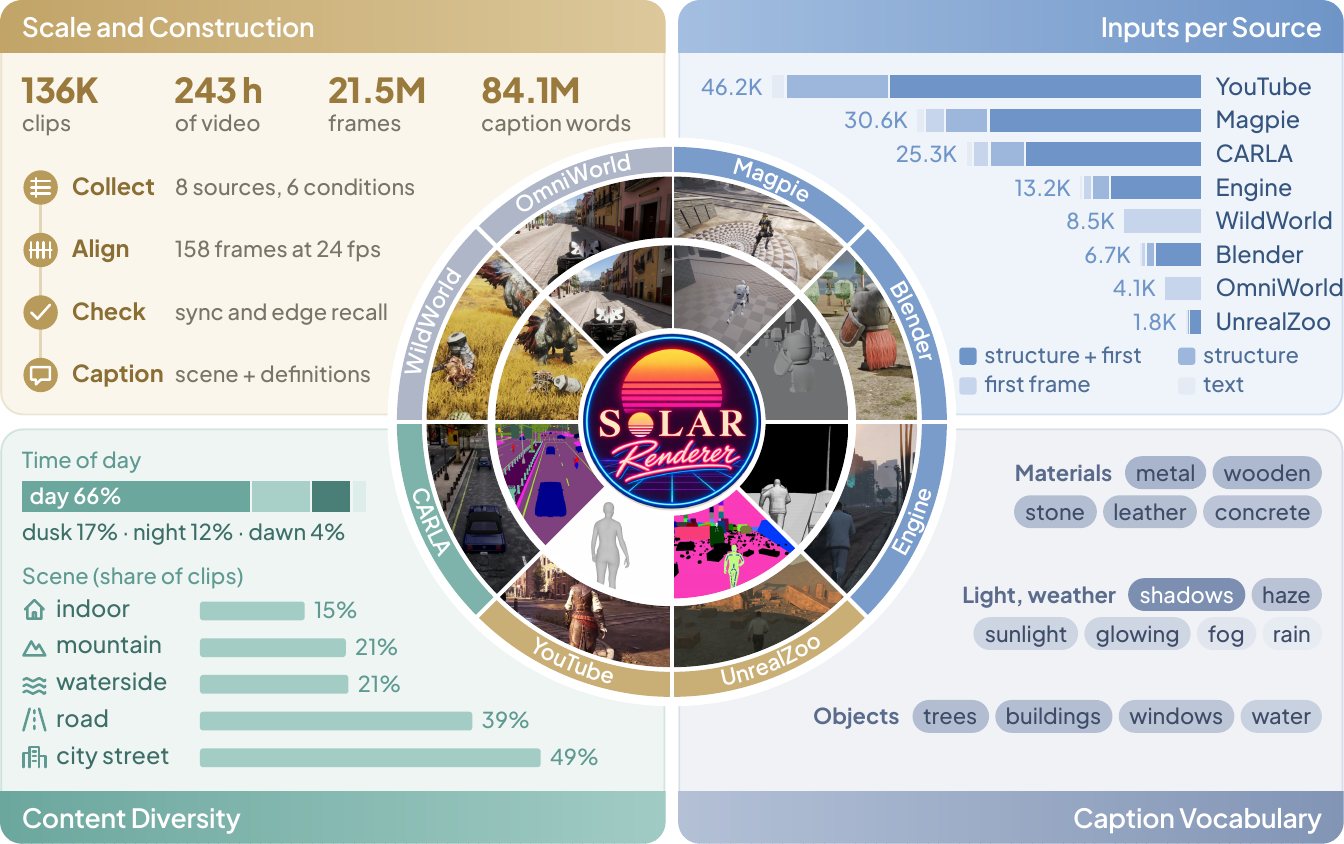}
  \caption{\textbf{The training corpus combines 136K clips from eight sources.} Each sector of the ring shows
  one source, with the condition on the inner ring (the first frame for
  WildWorld and OmniWorld) and the ground-truth frame of the same clip on
  the outer ring. The panels give the scale and construction steps of the
  data, the clips of each source by input combination, the time of day and
  scenes named in the captions, and the most frequent words of the
  captions across all training clips.}
  \label{fig:data}
\end{figure*}

\noindent\textbf{Structure-Guided Distribution Matching.}
The teacher is the fixed Stage-1 full-step model, and a separate LoRA
trained on student samples models the current student distribution. We
update $\psi$ with the distribution-matching gradient
\citep{yin2024dmd,yin2024dmd2}
\begin{equation}
 \nabla_{\psi}\mathcal{L}_{\mathrm{DMD}}
 =\mathbb{E}\!\left[
 \Bigl(\tfrac{\partial G_{\phi,\psi}}{\partial \psi}\Bigr)^{\!\top} w(t)
 \bigl(s_{\mathrm{fake}}(\widetilde z_t,t,\mathcal C)
       -s_{\mathrm{teacher}}(\widetilde z_t,t,\mathcal C)\bigr)
 \right],
 \label{eq:dmd}
\end{equation}
where $G_{\phi,\psi}$ is the four-step generator, $\mathcal C=(c,R(S,I))$,
and $\widetilde z_t$ is a re-noised student sample. Since full-step
sampling of the backbone is blurry, the composed student is already
sharper than its teacher, and plain distribution matching pulls it back
toward this blur: the adapted student ends below its own starting point
(\cref{sec:abl-guidance}). We therefore \textbf{\textit{amplify the
structural signal of the teacher}}. With $\hat z_0(\mathcal C)$ the clean
prediction of the teacher and $\hat z_0(\mathcal C_{\varnothing})$ its
prediction after the tokens of the structural video are removed from the
reference sequence, the teacher score is computed from
\begin{equation}
 \hat z_0^{\,w}
 =\hat z_0(\mathcal C)
 +(w-1)\bigl(\hat z_0(\mathcal C)-\hat z_0(\mathcal C_{\varnothing})\bigr).
 \label{eq:guidance}
\end{equation}
In $\mathcal C_{\varnothing}$ the first frame and the prompt are kept
unchanged. The difference isolates what the structural video contributes,
such as contours, layout, and character positions, so with $w>1$ the
student is pushed toward a target that is sharper and more decisive than
the teacher itself. Setting $w=1$ recovers plain distribution matching. The extrapolation
follows classifier-free guidance \citep{ho2022cfg}, which distribution
matching applies to the text condition of the teacher
\citep{yin2024dmd2,liu2026decoupleddmd}. Here it acts on the structural
condition alone.
The fake-score model is a separate rank-128 LoRA initialized at the
teacher, updated five times per generator update with learning rates of
$2\times10^{-6}$ and $10^{-5}$ for the fake-score model and the generator.
We set $w=1.5$ and run 800 iterations, each one fake-score update, at the
cost of one additional teacher forward pass per generator update
(\cref{app:implementation}).

\subsection{Causal Streaming Extension}
\label{sec:causal}

To generate video incrementally, we extend the same two-stage design to
block-causal attention. The clip is split into blocks of at most five
latent frames. We adapt the bidirectional control LoRA with teacher forcing
so that each block attends to the prompt, the first frame, the structure of
itself and the five preceding blocks, and the clean frames of the five
preceding blocks, never to a later block. We then freeze this causal
control LoRA and post-train a separate acceleration LoRA, initialized from
the bidirectional four-step LoRA, on the student's own block-by-block
rollouts, with the bidirectional control model as the structure-guided
teacher ($w=1.5$). Each rollout is generated without gradients and replayed
as one teacher-forced sequence, so that the gradient reaches both how each
block is generated and how it serves as context for later blocks. At
inference, we merge the acceleration LoRA into the backbone, keep the control LoRA
attached, sample each block in four steps, and update the fixed-window key--value cache before outputting the
block immediately. With a streaming VAE decoder~\citep{chen2026h3streaming} that
decodes each block without future latents, the causal model renders 768p
video in \textbf{\textit{real time}} at up to 28\,FPS.

\section{Engine-Aligned Paired Data}
\label{sec:data}

A renderer learns to follow structure only from pairs in which the target
video actually follows the structure. Conditions estimated from real
videos, such as predicted depth or segmentation, carry the errors of the
estimator and differ from the engine proxies that the renderer receives at
test time. We therefore collect pairs whose condition is exported by an
engine or fitted to the target, and use a pair as structural supervision
only when its condition matches the target in time and carries enough
structure. We then describe the data sources (\cref{sec:sources}), the
checks that decide how each source is used (\cref{sec:alignment}), and the
construction of prompts (\cref{sec:captions}).

\subsection{Data Sources}
\label{sec:sources}

The training set contains 136K samples, about 243 hours of video. Every
clip is a 158-frame window at 24\,fps and 768p resolution. The data cover
diverse sources and control types (\cref{fig:data}), listed below with
the condition each one supplies.

\begin{itemize}
  \item \textbf{Magpie} \citep{zhan2026magpie} records human-played
  Unreal Engine gameplay across many scenes at 60\,fps, together with a
  synchronized \textit{grey block-out} of the scene. The block-out gives the
  layout, the large geometry, the bodies and motion of the characters, and
  the camera of every frame, while materials and lighting are left to the
  model.
  \item \textbf{CARLA} \citep{dosovitskiy2017carla} provides urban and rural
  driving. Among the buffers that
  the simulator exports, we use \textit{semantic segmentation}, which paints
  every class in one flat colour and thus specifies the class layout of the
  scene without its appearance.
  \item \textbf{Engine recordings} consist of frames together with depth,
  object masks, camera poses, and skeletons on shared timestamps.
  We render the \textit{depth} as grey shading, with distant regions
  and the sky set to black, so that it conveys the shape of the scene near
  the camera.
  \item \textbf{Blender renders} of animated scenes are paired with the
  \textit{white-box} of their scene, which describes the scene layout and
  the motion of the main character.
  \item \textbf{UnrealZoo} \citep{zhong2024unrealzoo} provides Unreal
  Engine~5 environments in first- and third-person views. Its condition
  combines \textit{object masks and skeletons}, with one colour per object
  and the skeleton of each character drawn on top.
  \item \textbf{YouTube} videos \citep{ju2024miradata} provide clips for body-motion conditioning. We keep videos with a clear
  protagonist, fit the SMPL body \citep{loper2015smpl} of the protagonist with GVHMR
  \citep{shen2024gvhmr} using camera and depth from VGGT \citep{wang2025vggt} and Depth
Anything 3 \citep{lin2025depthanything3}, and render it from the target camera as a grey body on a
  white background. This \textit{SMPL body video} gives the position, pose,
  and motion of the protagonist, while the environment and clothing come
  from the first frame and text.
  \item \textbf{WildWorld} \citep{li2026wildworld}, a public dataset of
  \textit{Monster Hunter Wilds} gameplay, and \textbf{OmniWorld}
  \citep{zhou2025omniworld}, which collects footage of AAA games such as
  \textit{Cyberpunk 2077}, contribute rich game appearance. Their
  structural annotations follow the frames less closely, so we use them as
  appearance data and supply only a \textit{first frame}.
\end{itemize}

\begin{table}[t]
\centering
\caption{\textbf{Comparison with existing methods.} Gemini Elo ratings
for the four judge questions and the automatic score (Auto.) on both test
sets. The causal variant is rated only against ours. Methods that lose
every comparison on a question share one rating. Bold and underlined
italics mark the best and second-best values.}
\label{tab:main}
\tablefont
\setlength{\tabcolsep}{1pt}
\renewcommand{\arraystretch}{1.2}
\begin{tabularx}{\linewidth}{l c*{10}{>{\centering\arraybackslash}X}}
\toprule
\rowcolor{tabhead}
& & \multicolumn{5}{c}{{\fontsize{7.5}{9}\selectfont\itshape Generalization benchmark}} & \multicolumn{5}{c}{{\fontsize{7.5}{9}\selectfont\itshape In-distribution test set}}\\
\rowcolor{tabhead}
\resulthead{Method} & \resulthead{Steps} & \resulthead{Overall} & \resulthead{Quality} & \resulthead{Structure} & \resulthead{First fr.} & \resulthead{Auto.} & \resulthead{Overall} & \resulthead{Quality} & \resulthead{Structure} & \resulthead{First fr.} & \resulthead{Auto.}\\
\midrule
\rowcolor{tabgroup}
\multicolumn{12}{l}{\textbf{Baselines}}\\
Official backbone~\citeyearpar{minimax2026h3} & $50$ & $1075$ & $1086$ & $627$ & $1153$ & $69.8$ & $651$ & $586$ & $554$ & $660$& $64.8$\\
\rowcolor{tabstripe}
\textbf{Ours (Stage~1)} & $50$ & \underline{$\mathit{1389}$} & $1339$ & \underline{$\mathit{1276}$} & $1395$ & $\mathbf{71.6}$ & $1299$ & $1196$ & $1308$ & $1346$& $66.2$\\
\midrule
\rowcolor{tabgroup}
\multicolumn{12}{l}{\textbf{Structure-conditioned renderers}}\\
Wan2.2-VACE-Fun~\citeyearpar{jiang2025vace} & $50$ & $389$ & $489$ & $763$ & $377$ & $65.2$ & $651$ & $586$ & $554$ & $660$& $63.0$\\
\rowcolor{tabstripe}
PAI ControlNet-Union~\citeyearpar{pai2026controlnetunion} & $40$ & $1221$ & $1228$ & $1083$ & $1150$ & $69.4$ & $1189$ & $1304$ & $1187$ & $1029$& $64.6$\\
Cosmos-Transfer2.5~\citeyearpar{nvidia2025cosmos} & $35$ & $389$ & $368$ & $1007$ & $377$ & $64.7$ & $651$ & $991$ & $1012$ & $660$& $64.7$\\
\rowcolor{tabstripe}
LynnReal-Omni~\citeyearpar{mao2026lynnrealomninativemultimodalvideo} & $4$ & $\mathbf{1441}$ & \underline{$\mathit{1412}$} & $1241$ & \underline{$\mathit{1423}$} & $70.0$ & \underline{$\mathit{1410}$} & \underline{$\mathit{1359}$} & $1295$ & \underline{$\mathit{1362}$}& $64.2$\\
Streaming CWM~\citeyearpar{liao2026streamingcwm} & $4$ & $1075$ & $1100$ & $930$ & $1199$ & $70.7$ & $988$ & $968$ & $955$ & $1102$& $64.2$\\
\rowcolor{tabstripe}
StreamRender-H3~\citeyearpar{nvidia2026streamrender} & $2$ & $580$ & $560$ & $737$ & $498$ & $63.9$ & $651$ & $586$ & $750$ & $660$& $59.0$\\
\midrule
\textbf{Ours (causal)} & $4$ & $1191$ & $1145$ & $1235$ & $1186$ & \underline{$\mathit{71.2}$} & $1199$ & $1120$ & \underline{$\mathit{1313}$} & $1235$& \underline{$\mathit{66.6}$}\\
\rowcolor{tabours}
\textbf{Ours} & $4$ & $\mathbf{1441}$ & $\mathbf{1420}$ & $\mathbf{1338}$ & $\mathbf{1428}$ & $\mathbf{71.6}$ & $\mathbf{1510}$ & $\mathbf{1424}$ & $\mathbf{1386}$ & $\mathbf{1522}$& $\mathbf{67.0}$\\
\bottomrule
\end{tabularx}
\end{table}

\subsection{Data Checks}
\label{sec:alignment}

A condition can fail in two ways. It can be offset from the target in time
or space, or it can be aligned yet too coarse to describe the target. We
therefore run two checks on every source.

\noindent\textbf{Temporal and Spatial Agreement.}
For timing, we correlate the motion of each condition with that of its
target over a range of frame offsets. The offset at which the correlation
peaks shows whether the two are synchronized, and clips whose peak is not
at zero offset are removed. For space, we measure the recall of the
condition edges in the target frame and compare it with the recall of the
horizontally mirrored condition. A condition that matches its target
recovers clearly more edges than its mirror, so clips without this margin
are removed as spatially misaligned. Further filters remove near-duplicates,
dark or low-contrast clips, and clips whose condition covers too little of
the frame. Finally, we manually inspect samples from every source before
it enters training.

\noindent\textbf{Sufficiency of the Condition.}
For the data, we first used box proxies that draw body parts and objects as
coarse boxes. Although they are synchronized with the frames, the boxes
omit most of the scene geometry, so a model trained on them cannot learn
the shapes it is asked to follow. We therefore switched to depth
within 30\,m rendered as grey shading, with farther regions and the sky set
to black, which carries the layout and object shapes of the scene without
the strong flicker of finer block proxies. For the same reason, WildWorld
boxes, which do not line up with the frames, and OmniWorld background
blocks, which omit people and moving objects, are not used as structure.

\subsection{Prompts and Input Combinations}
\label{sec:captions}

\noindent\textbf{Captions.}
Each caption follows the six-section prompt format of the backbone, which
consists of subject definitions, a summary with a task prefix, a
retention analysis, a detailed description, the soundscape, and the music
(\cref{fig:pipeline}). Two parts fill these sections. \sn{1}A
\textit{scene description} covers materials, colours, illumination,
weather, character appearance, events, camera motion, and sound, and
fills the summary, the detailed description, the soundscape, and the
music. Kimi K2.6 \citep{moonshot2026kimik26,kimi2026k25} writes it after watching only the target video. \sn{2}A \textit{fixed
part} defines each reference in the subject definitions, names the task
in the summary prefix, and states in the retention analysis which
attributes of each reference the output follows. Code generates it from
the input combination of the clip (\cref{sec:prompts}), so the
description of each condition stays identical across clips.

\begin{figure*}[t]
\centering
\includegraphics[width=\linewidth]{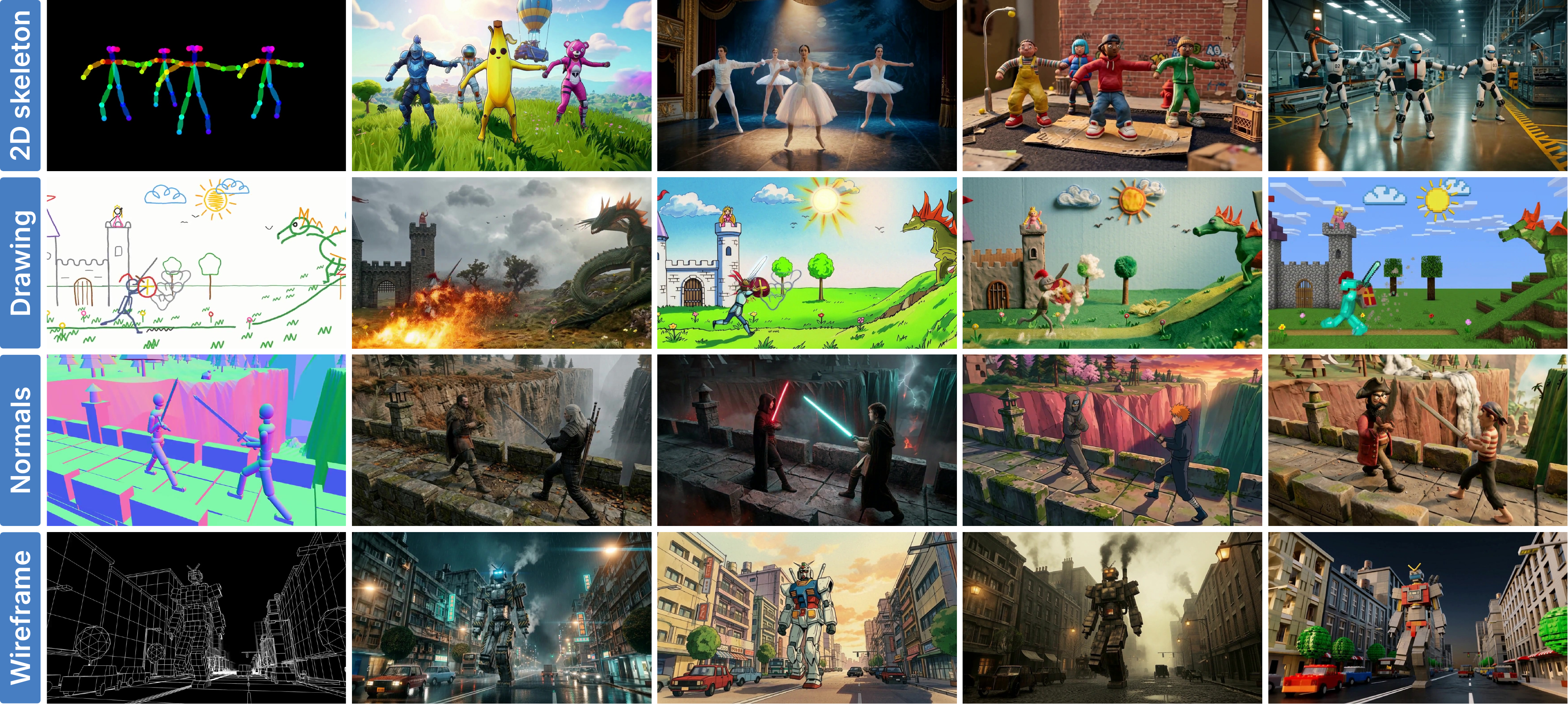}
\caption{\textbf{Multiple appearances from unseen structural representations.}
The left column shows, from top to bottom, a 2D skeleton, a crayon drawing,
a normal map, and a wireframe. The remaining columns show our generated
appearances at the same time.}
\label{fig:qualitative-styles}
\end{figure*}

\noindent\textbf{Input Combinations.}
The training samples mix four input combinations, with 67.1\% supplying
structure and a first frame, 16.7\% structure only, 13.2\% a first frame
only, and 2.9\% text only. Sources with structure also contribute some
samples without it, so that missing structure is not tied to the
appearance of WildWorld and OmniWorld. The combination of each clip is
fixed when the data is built, and its prompt describes exactly the
supplied inputs. Dropping the
structure at random during training would leave a prompt that still refers
to it, so fixing the combination in advance \textbf{\textit{prevents the
absent input from leaking through the prompt}}.

\section{Experiments}
\label{sec:experiments}

We first describe the evaluation setups(\cref{sec:setup}), then compare \project with existing
structure-conditioned renderers and the baselines
(\cref{sec:main-results}), and finally present ablations on the structure-guided
post-training with existing acceleration LoRAs, the
conditioning design, and the training data (\cref{sec:ablations}).

\begin{figure*}[t]
\centering
\includegraphics[width=\linewidth]{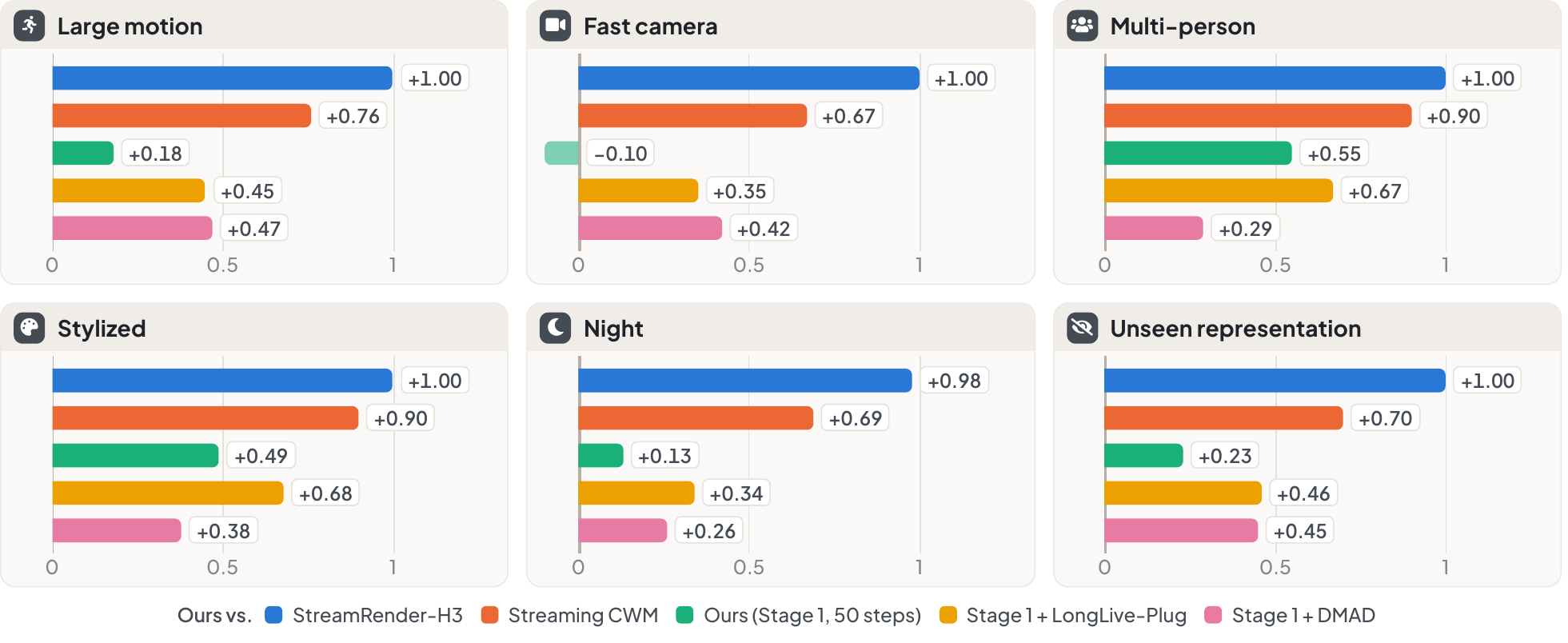}
\caption{\textbf{Results by difficulty tag.} Net score of ours against each
compared method on the generalization benchmark. Positive values favor ours.}
\label{fig:by-tag}
\end{figure*}

\subsection{Experimental Setup}
\label{sec:setup}
\label{sec:evaluation}

\noindent\textbf{Evaluation Sets.}
We evaluate on two test sets, each answering a different question.
\begin{itemize}
  \item \textbf{In-distribution test set}, which assesses how well \project
  renders the kinds of structure seen in training. Its 53 clips come from
  the training sources, share no recording, scene, or map with the training data, and each has a ground-truth video.
  \item \textbf{Generalization benchmark}, which assesses how well a single
  model transfers to new structure and has no ground truth. Its 208 cases
  are built from 57 structural videos covering 16 representations, 11 of
  which \textbf{\textit{never occur in training}}, such as pencil
  storyboards, children's drawings, point clouds, voxels, and wireframes. Most of these videos were built in a game engine by
  a coding agent, and each case is labelled by difficulty, namely large
  motion, fast camera motion, multiple people, stylization, and night
  scenes, with results per label in \cref{fig:by-tag}.
\end{itemize}

\begin{figure*}[t]
\centering
\includegraphics[width=\linewidth]{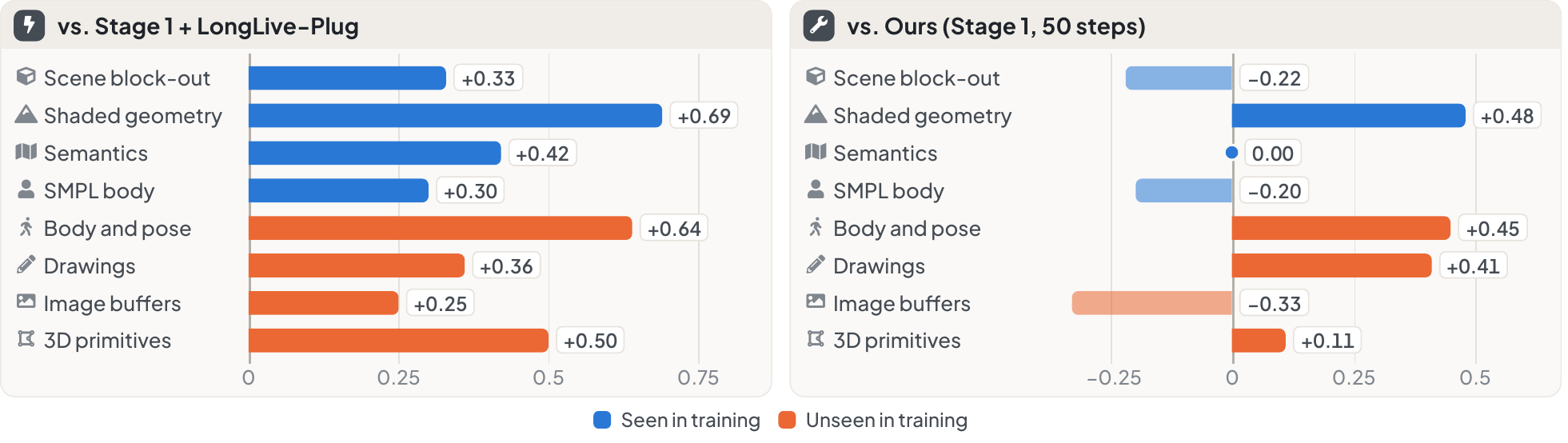}
\caption{\textbf{Results by representation.} Net score of ours against
Stage~1 + LongLive-Plug (left) and our 50-step Stage~1 model (right) on the
generalization benchmark, grouped by representation category. Positive
values favor ours.}
\label{fig:by-repr}
\end{figure*}

\noindent\textbf{Evaluation Protocol.}
We assess every method with pairwise judgments and automatic scores.
\begin{itemize}
  \item \textbf{Pairwise judgment} is our primary measure. Gemini~3.8 Flash \citep{google2026gemini38flash,team2023gemini}
  receives the structural video, the first frame, and two renderings, and
  compares them on \textsuperscript{1}\textit{visual quality},
  \textsuperscript{2}\textit{adherence to the structure and motion},
  \textsuperscript{3}\textit{preservation of the first frame}, and
  \textsuperscript{4}\textit{overall preference}. Each pair is judged in
  both orders. We report the \textbf{net score} of ours against each
  compared method as (wins $-$ losses)/cases, the difference between our
  win and loss rates. It ranges from $-1$ to $1$, with positive values
  indicating a preference for ours. From the verdicts between all pairs of methods on a test
  set we fit one Bradley--Terry model \citep{bradley1952rank} per judge
  question and report the strength of every method as an \textbf{Elo
  rating}, with the mean of the compared methods at 1000. A blind human study agrees
  with the verdicts of the judge.
  \item \textbf{Automatic metrics}, grouped into
  \textsuperscript{1}\textit{quality} (DOVER \citep{wu2023dover}, LAION
  aesthetics \citep{schuhmann2022aesthetic}, VideoScore2
  \citep{he2025videoscore2}), \textsuperscript{2}\textit{stability}
  (flow-warping residual and adjacent-frame DINOv2 similarity
  \citep{oquab2023dinov2}), \textsuperscript{3}\textit{control} (CoTracker3
  track error \citep{karaev2024cotracker3} and edge F1 against the
  structure), and \textsuperscript{4}\textit{first-frame preservation}
  (DINOv2 similarity of the last second to the first frame). Each metric
  is mapped linearly from a fixed range to 0--100, and the
  \textbf{automatic score} averages the metrics within each group and then
  the four groups.
\end{itemize}
Details of the protocol and the full results are given in
\cref{app:evaluation,app:results}.

\begin{figure*}[t]
\centering
\includegraphics[width=\linewidth]{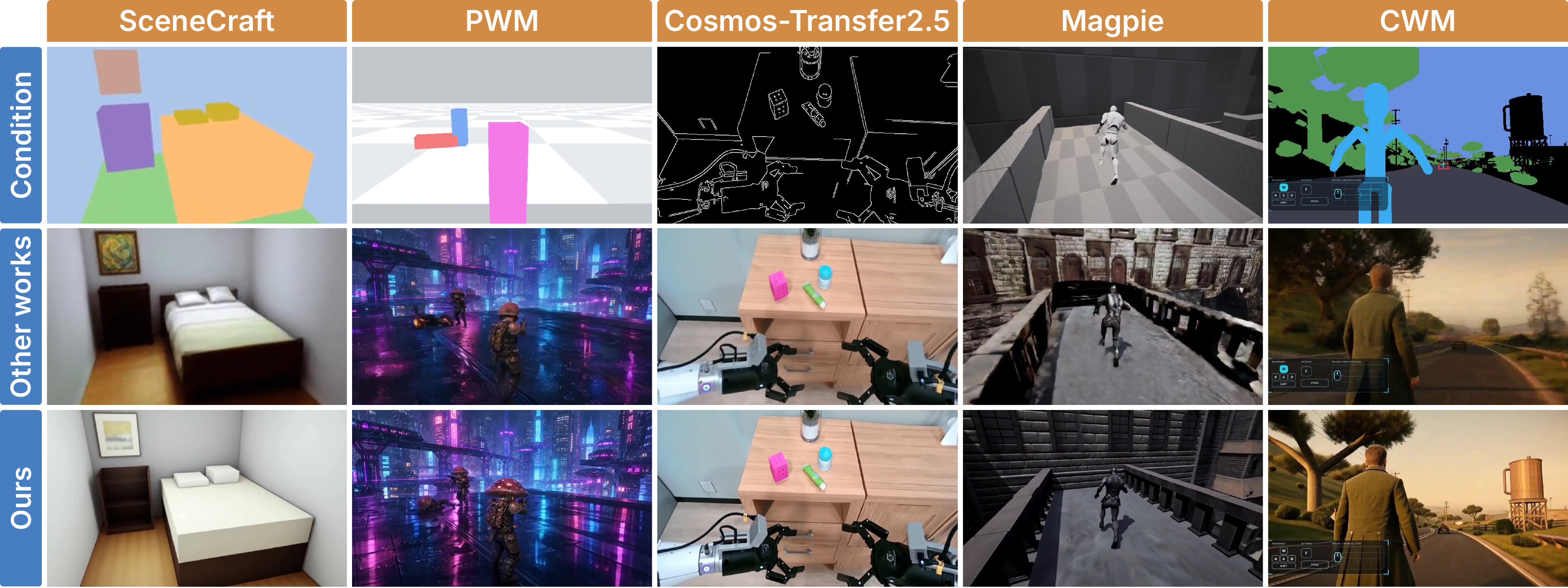}
\caption{\textbf{Visual comparison with other methods.}
Top to bottom: structural condition, results from other works, and ours.
Left to right: SceneCraft, Programmable World Model, Cosmos-Transfer2.5,
Magpie, and CWM, with frames synchronized within each example.}
\label{fig:qualitative}
\end{figure*}

\noindent\textbf{Compared Methods.}
We compare \project with three groups of methods.
\begin{itemize}
  \item \textbf{Baselines}, the official Ref2VA backbone and our Stage~1 model, both with 50 steps.
  \item \textbf{Existing acceleration LoRAs}, which assess the
  structure-guided post-training of Stage~2. Each attaches LongLive-Plug
  \citep{longlive2026plug}, DMAD \citep{yu2026dmad}, or FastH3
  \citep{fastvideo2026fasth3} to our Stage~1 model without further training.
  \item \textbf{Structure-conditioned renderers}, namely LynnReal-Omni
  \citep{mao2026lynnrealomninativemultimodalvideo}, StreamRender-H3
  \citep{nvidia2026streamrender}, Streaming CWM \citep{liao2026streamingcwm},
  PAI ControlNet-Union \citep{pai2026controlnetunion}, Cosmos-Transfer2.5
  \citep{nvidia2025cosmos}, and Wan2.2-VACE-Fun
  \citep{wan2025wan,jiang2025vace}. Each runs with its released resolution,
  frame count, sampler, and number of steps, and its outputs are resized to
  our evaluation format.
\end{itemize}

\noindent\textbf{Implementation.}
Training builds on the open SolarWM framework~\citep{huang2026solarwm}. \project uses the control LoRA after 38,400 training steps. Its
acceleration LoRA has rank 384, with the first 128 ranks initialized from
the pretrained DMAD four-step LoRA and the rest at zero, and is trained for
800 iterations of structure-guided distribution matching with $w=1.5$.
It renders a 158-frame clip with four denoising steps in about 209\,s on
one GPU, 12.3 times faster than the 2,562\,s of 50-step sampling.

\subsection{Comparison with Existing Methods}
\label{sec:main-results}

\begin{table*}[t]
\begin{minipage}[t]{0.485\linewidth}
\centering
\caption{\textbf{Ablation on structure guidance.} Gemini Elo
ratings on the generalization benchmark, fitted per starting point and
comparable only within a block. $w=1$ means no guidance, and bold marks
the best value per block.}
\label{tab:abl-guidance}
\tablefont
\setlength{\tabcolsep}{2pt}
\renewcommand{\arraystretch}{1.2}
\begin{tabularx}{\linewidth}{l*{4}{>{\centering\arraybackslash}X}}
\toprule
\rowcolor{tabhead}
\resulthead{Model} & \resulthead{Overall} & \resulthead{Quality} & \resulthead{Struct.} & \resulthead{First fr.}\\
\midrule
\rowcolor{tabgroup}
\multicolumn{5}{l}{\textbf{Self-distillation}}\\
\quad Start (LongLive-Plug) & $986$ & $985$ & $984$ & $\mathbf{999}$\\
\quad $w=1$ & $949$ & $947$ & $979$ & $981$\\
\quad $w=2$ & $\mathbf{1066}$ & $\mathbf{1069}$ & $\mathbf{1037}$ & $1020$\\
\midrule
\rowcolor{tabgroup}
\multicolumn{5}{l}{\textbf{LongLive-Plug start}}\\
\quad No post-training & $1028$ & $1033$ & $997$ & $1028$\\
\quad $w=1$ & $798$ & $789$ & $980$ & $856$\\
\quad $w=1.5$ & $1067$ & $1065$ & $\mathbf{1012}$ & $1046$\\
\quad $w=2$ & $\mathbf{1107}$ & $\mathbf{1113}$ & $\mathbf{1012}$ & $\mathbf{1070}$\\
\midrule
\rowcolor{tabgroup}
\multicolumn{5}{l}{\textbf{DMAD start}}\\
\quad No post-training & $1024$ & $1027$ & $1003$ & $1040$\\
\quad $w=1$ & $820$ & $812$ & $982$ & $850$\\
\rowcolor{tabours}
\quad \textbf{Ours ($\boldsymbol{w=1.5}$)} & $\mathbf{1156}$ & $\mathbf{1161}$ & $\mathbf{1015}$ & $\mathbf{1110}$\\
\bottomrule
\end{tabularx}
\end{minipage}
\hfill
\begin{minipage}[t]{0.485\linewidth}
\centering
\caption{\textbf{Ablation on post-training settings.} Gemini Elo
ratings on the generalization benchmark from one fit over this table,
comparable only within it. Each block varies one setting of our model,
listed in the last row.}
\label{tab:abl-design-main}
\tablefont
\setlength{\tabcolsep}{2pt}
\renewcommand{\arraystretch}{1.477}
\begin{tabularx}{\linewidth}{l*{4}{>{\centering\arraybackslash}X}}
\toprule
\rowcolor{tabhead}
\resulthead{Setting} & \resulthead{Overall} & \resulthead{Quality} & \resulthead{Struct.} & \resulthead{First fr.}\\
\midrule
\rowcolor{tabgroup}
\multicolumn{5}{l}{\textbf{Starting point}}\\
\quad LongLive-Plug & $922$ & $910$ & $993$ & $956$\\
\quad FastH3 & $1053$ & $1061$ & $1000$ & $1049$\\
\quad DMAD & $1006$ & $1011$ & $992$ & $1012$\\
\midrule
\rowcolor{tabgroup}
\multicolumn{5}{l}{\textbf{Guidance strength}}\\
\quad $w=2$ & $1045$ & $1062$ & $996$ & $1059$\\
\midrule
\rowcolor{tabgroup}
\multicolumn{5}{l}{\textbf{Sampling steps}}\\
\quad 2 steps & $961$ & $969$ & $970$ & $983$\\
\quad 8 steps & $1080$ & $1094$ & $\mathbf{1010}$ & $1032$\\
\midrule
\rowcolor{tabours}
\textbf{Ours (DMAD, $\boldsymbol{w=1.5}$, 4 steps)} & $\mathbf{1096}$ & $\mathbf{1108}$ & $1005$ & $\mathbf{1076}$\\
\bottomrule
\end{tabularx}
\end{minipage}

\end{table*}

\noindent\textbf{Comparison with Structure-Conditioned Renderers.}
\project reaches the \textbf{\textit{best rating on every judge question of
both test sets}} (\cref{tab:main}). Renderers built around one engine do
not transfer. StreamRender-H3 and Streaming CWM learned a single
representation from a racing game and from CARLA, and rate 580 and 1075
overall on the generalization benchmark against 1441 for ours.
Cosmos-Transfer2.5 and Wan2.2-VACE-Fun trace the edges of the structure
closely, yet their appearance falls so far behind that they rate below
400. The strongest competitor, LynnReal-Omni, ties with ours in overall
preference on the generalization benchmark (1441 each) and trails by 100
points on the in-distribution test set, while its structure rating is
about 95 points lower on both. LynnReal-Omni labels the type of each reference, whereas our method reads untyped,
frame-aligned structure, which \textbf{\textit{tightens its control}}.
Its total automatic score also exceeds every external renderer on both sets,
with balanced quality and structural agreement.

\noindent\textbf{Causal Streaming.}
The causal variant of our method renders each block from past frames only
and outputs it as soon as it is sampled. It rates 1191 and 1199 overall on
the two test sets, below the bidirectional model at 1441 and 1510 but well
above the streaming renderers, Streaming CWM at 1075 and 988 and
StreamRender-H3 at 580 and 651 (\cref{tab:main}). Its structure rating
stays close to ours and on the in-distribution test set even exceeds our
Stage~1 model (1313 against 1308), so that causal attention mainly costs
appearance, as later blocks drift from the colours of the first frame,
while \textbf{\textit{the learned control carries over}} to block-by-block
generation.

\noindent\textbf{Generalization on Hard Cases.}
\Cref{fig:by-tag} splits the generalization benchmark by difficulty tag.
Ours is preferred over every compared method on every tag, including large
motion, multiple people, and night scenes, with the single exception of
our own 50-step model on fast camera motion ($-0.10$). The margin
over the engine-specific renderers stays between $+0.67$ and $+1.00$, and
the margin over the four-step compositions with LongLive-Plug and DMAD
between $+0.26$ and $+0.68$. The cases with an unseen representation are
no exception, with $+0.46$ over LongLive-Plug and $+0.70$ over Streaming
CWM, and ours also keeps the best structure rating among all methods on
them (\cref{tab:unseen}).

\noindent\textbf{Generalization to Unseen Representations.}
\Cref{fig:by-repr} groups the representations into categories. Ours is
ahead of LongLive-Plug on every category, seen or unseen, from block-outs
and semantic maps to pose skeletons, drawings, image buffers, and 3D
primitives, by $+0.25$ to $+0.69$. Against our own 50-step model, the
four-step model is preferred on body and pose ($+0.45$) and drawings
($+0.41$), so acceleration costs little on unseen structure. Since no
reference carries a type identifier, the control LoRA cannot specialize
to a format and instead learns to read layout and motion from whatever
occupies the position of a target frame. A single control LoRA therefore
\textbf{\textit{generalizes to unseen structure}}.

\begin{table*}[t]
\begin{minipage}[t]{0.485\linewidth}
\centering
\caption{\textbf{Ablation on conditioning design.} Automatic
group scores and their mean (0--100) for control models. Bold marks the best per block.}
\label{tab:injection}
\tablefont
\renewcommand{\resulthead}[1]{{\fontsize{6.5}{8}\selectfont\sffamily\bfseries\shortstack{#1}}}
\setlength{\tabcolsep}{1.2pt}
\renewcommand{\arraystretch}{1.2}
\begin{tabularx}{\linewidth}{l*{5}{>{\centering\arraybackslash}X}}
\toprule
\rowcolor{tabhead}
\resulthead{Model} & \resulthead{Qual.} & \resulthead{Stab.} & \resulthead{Ctrl.} & \resulthead{First} & \resulthead{Auto.}\\
\midrule
\rowcolor{tabgroup}
\multicolumn{6}{l}{\textbf{Conditioning pathway}}\\
ControlNet branch & $\mathbf{60.2}$ & $90.5$ & $46.8$ & $66.4$ & $66.0$\\
\rowcolor{tabours}
\textbf{Ours (native)} & $59.6$ & $\mathbf{91.6}$ & $\mathbf{48.0}$ & $\mathbf{66.8}$ & $\mathbf{66.5}$\\
\midrule
\rowcolor{tabgroup}
\multicolumn{6}{l}{\textbf{Structure resolution}}\\
Half resolution & $59.5$ & $91.5$ & $46.6$ & $\mathbf{67.0}$ & $66.2$\\
\rowcolor{tabours}
\textbf{Ours (full res.)} & $\mathbf{59.6}$ & $\mathbf{91.6}$ & $\mathbf{48.0}$ & $66.8$ & $\mathbf{66.5}$\\
\midrule
\rowcolor{tabgroup}
\multicolumn{6}{l}{\textbf{RoPE (half resolution)}}\\
Separate RoPE & $59.1$ & $\mathbf{91.5}$ & $46.4$ & $\mathbf{67.0}$ & $66.0$\\
Shared RoPE & $\mathbf{59.5}$ & $\mathbf{91.5}$ & $\mathbf{46.6}$ & $\mathbf{67.0}$ & $\mathbf{66.2}$\\
\bottomrule
\end{tabularx}
\end{minipage}\hfill
\begin{minipage}[t]{0.485\linewidth}
\centering
\caption{\textbf{Ablation on training data.} Automatic group scores
and their mean (0--100) for data variants. Bold marks the best per block.}
\label{tab:abl-data}
\tablefont
\renewcommand{\resulthead}[1]{{\fontsize{6.5}{8}\selectfont\sffamily\bfseries\shortstack{#1}}}
\setlength{\tabcolsep}{1.2pt}
\renewcommand{\arraystretch}{1.4}
\begin{tabularx}{\linewidth}{l*{5}{>{\centering\arraybackslash}X}}
\toprule
\rowcolor{tabhead}
\resulthead{Model} & \resulthead{Qual.} & \resulthead{Stab.} & \resulthead{Ctrl.} & \resulthead{First} & \resulthead{Auto.}\\
\midrule
\rowcolor{tabgroup}
\multicolumn{6}{l}{\textbf{Data composition}}\\
All data & $59.5$ & $91.5$ & $46.6$ & $\mathbf{67.0}$ & $66.2$\\
Structured only & $59.7$ & $91.2$ & $47.3$ & $66.8$ & $66.2$\\
\rowcolor{tabours}
\textbf{Ours (w/o misaligned)} & $\mathbf{59.9}$ & $\mathbf{91.6}$ & $\mathbf{47.4}$ & $66.9$ & $\mathbf{66.5}$\\
\midrule
\rowcolor{tabgroup}
\multicolumn{6}{l}{\textbf{Data scale}}\\
$1/10$ data & $\mathbf{59.6}$ & $91.4$ & $46.5$ & $\mathbf{67.1}$ & $66.1$\\
Full data, new seed & $58.9$ & $\mathbf{91.5}$ & $46.4$ & $66.9$ & $65.9$\\
All data & $59.5$ & $\mathbf{91.5}$ & $\mathbf{46.6}$ & $67.0$ & $\mathbf{66.2}$\\
\bottomrule
\end{tabularx}
\end{minipage}
\end{table*}

\noindent\textbf{Comparison with the Backbone.}
Without the control LoRA, the official backbone treats the structural
video as a generic reference and rarely follows its layout. Its structure
rating is 627 on the generalization benchmark and 554 on the
in-distribution test set, against 1338 and 1386 for ours
(\cref{tab:main}), so the control comes entirely from a lightweight LoRA
on top of a frozen backbone. Our 50-step Stage~1 model already follows the
structure, which shows that the native reference pathway with a single
control LoRA suffices for control, but it renders blurrier frames. The
four-step final model rates higher on both test sets, 1441 against 1389
and 1510 against 1299 overall, and renders 12.3 times faster.

\noindent\textbf{Visual Comparison with Other Methods.}
\Cref{fig:qualitative} compares our four-step outputs with results from
five renderers, using the structural inputs from their demonstration cases. Ours produces sharper furniture boundaries in
SceneCraft~\citep{yang2024scenecraft} and clearer character silhouettes
in the neon scene from Programmable World Model. In the Magpie scene, the bridge railings follow the
block-out more closely, while the CWM example preserves the roadside layout
and resolves the water tower. The Cosmos example preserves the tabletop layout from edge input.

\begin{figure*}[t]
\centering
\includegraphics[width=\linewidth,height=0.87\textheight,keepaspectratio]{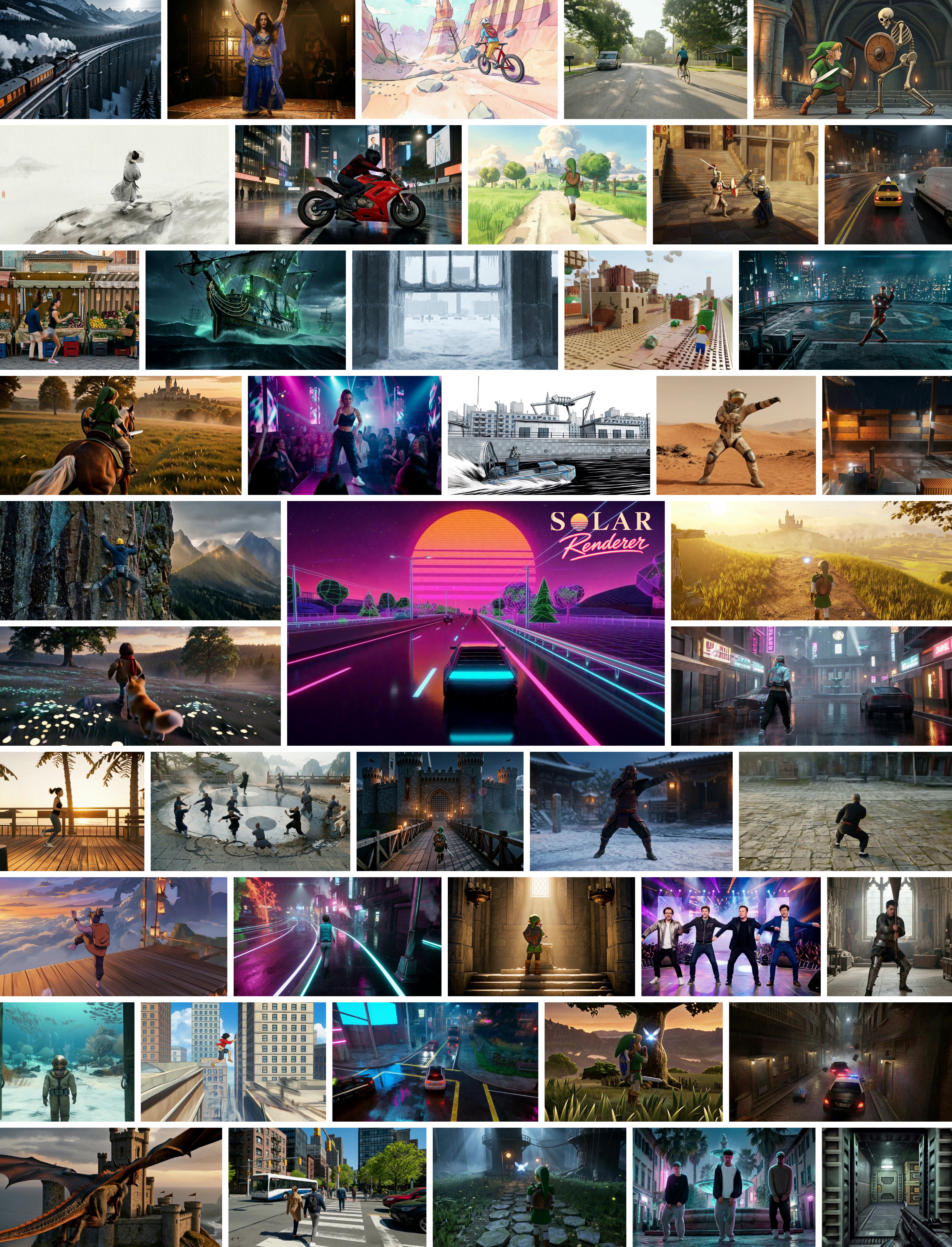}
\caption{\textbf{A gallery of structure-conditioned renderings.}
Each image comes from a different case, spanning human motion, everyday scenes,
and imagined environments across seen and unseen structural representations.}
\label{fig:result-gallery}
\end{figure*}

\noindent\textbf{Visual Diversity.}
\Cref{fig:qualitative-styles} illustrates how the same unseen structure supports
contrasting appearances. The dancers preserve their arrangement across human,
clay, and robotic forms, while the drawing, normal map, and wireframe examples
change materials and scene style around the specified composition.
\Cref{fig:result-gallery} presents additional showcase renderings across scenes
and visual styles.

\subsection{Ablation Studies}
\label{sec:ablations}

We ablate the post-training stage on the generalization benchmark with
Gemini Elo ratings (\cref{tab:abl-guidance,tab:abl-design-main}), and the
conditioning design and the training data scored by automatic metrics
(\cref{tab:injection,tab:abl-data}).

\noindent\textbf{Comparison with Existing Acceleration LoRAs.}
An existing four-step LoRA attached to our control LoRA renders in four
steps, but it was trained without the control adapter and shifts the
outputs of the control model. Composed this way, LongLive-Plug, DMAD, and
FastH3 reach overall ratings of 922, 1006, and 1053
(\cref{tab:abl-design-main}). Structure-guided post-training removes the
mismatch and lifts the DMAD start to 1096, above every composition, with
the highest quality and first-frame ratings. The gain therefore comes from
adapting the acceleration LoRA to the control model.

\noindent\textbf{Structure Guidance.}
\label{sec:abl-guidance}
\Cref{tab:abl-guidance} post-trains the same students without guidance
($w=1$), either by updating the four-step LoRA from DMAD or LongLive-Plug,
or by self-distillation, which updates the control LoRA with LongLive-Plug
frozen. Plain distribution matching pulls the student toward a teacher
whose structure-conditioned prediction is blurry, so the unguided student
\textbf{\textit{degrades its starting point}}, from 1024 to 820 for DMAD and
from 1028 to 798 for LongLive-Plug. Extrapolating the teacher along the
structural condition sharpens this target, and the guided student rates
far above both the unguided student and its start from every starting point, 1156 against 820
from DMAD, 1107 against 798 from LongLive-Plug, and 1066 against 949 with
self-distillation, with the best value on almost every judge question.
A blind human comparison with the unguided DMAD student
agrees in direction.

\noindent\textbf{Starting Point and Strength.}
Guidance improves every starting point, and DMAD with $w=1.5$ gives the
best student. The best strength depends on the start. From LongLive-Plug,
$w=2$ is better than $w=1.5$ (1107 against 1067 in
\cref{tab:abl-guidance}), whereas from DMAD, $w=2$ lowers the rating from 1096 to 1045 (\cref{tab:abl-design-main}).

\noindent\textbf{Schedule, Training Length, and Sampling Steps.}
Applying guidance only at re-noising levels $\sigma\ge0.9$ rates on par
with guidance at every level (1107 against 1096), so guidance acts mainly
at high noise levels, where the layout of the clip is decided. Longer
post-training does not help, as 1,200 iterations lower the overall rating to
992, and adding self-distillation of the control LoRA on top of our
post-training lowers it to 1049. The student is trained for four
steps and works best with four. Two steps lower the overall rating to 961,
and eight steps bring no gain (1080).

\noindent\textbf{Conditioning Design.}
\label{sec:abl-conditioning}
\Cref{tab:injection} trains variants of the control model under the same
data and budget. A ControlNet branch in place
of the native reference pathway lowers the structure score from 48.0 to
46.8 and the stability, and giving the structure tokens separate RoPE
positions instead of those of their target frames lowers quality and
structure, which supports \textbf{\textit{native reference conditioning}}.
Reading the structure at full instead of half resolution further raises
the structure score from 46.6 to 48.0.

\noindent\textbf{Frame-Aligned Data.}
\label{sec:abl-data}
\Cref{tab:abl-data} varies the training data. Using
the WildWorld boxes, which do not line up with the frames, as structure
lowers the structure score from 47.4 to 46.6, since misaligned supervision
teaches the model to loosen its control. Training only on samples
with structure lowers stability. The final corpus therefore keeps samples
without structure for appearance and uses as structure only sources whose structure is
\textbf{\textit{aligned frame by frame}} with the video. Training on $1/10$
of the data changes every score by at most 0.2, less than retraining on
the full data with a new seed, whereas removing the misaligned structure
helps more, so the alignment of the data matters more than its amount.

\section{Conclusion}
\label{sec:conclusion}

We presented \project, a few-step generative renderer that turns structure
from game engines and coding agents into realistic video. Structural videos
enter the native reference pathway of a frozen MiniMax-H3 backbone without
any type identifier, so that a single control LoRA follows arbitrary
structural representations, and structure-guided distribution matching
adapts a pretrained four-step LoRA to this control model. Trained on 136K
frame-aligned clips from game engines, simulators, and YouTube videos, our method achieves state-of-the-art results on both the in-distribution test
set and the generalization benchmark, and generalizes to representations
never seen in training, while rendering 12.3 times faster than 50-step sampling. Its block-causal
variant further renders 768p video in \textbf{\textit{real time}}.

\clearpage
\solarappendix
\section{Experimental Details}
\label{app:details}

\begin{figure}[t]
\centering
\begin{promptcard}
\promptfield{subject\_definitions}<Video~1> is a semantic segmentation of the scene, one flat colour per class (road, sidewalk, building, vegetation, vehicle, pedestrian, sky and so on), frame by frame, from the same camera as the target video.\\
<Video~2> is the first frame of [Shot 1], held still.\par\medskip
\promptfield{summary}[keyframe completion + reference generation] The target video begins from <Video~2> and follows the frame-by-frame structure and motion of <Video~1>. A blue hatchback drives along a foggy, rain-slicked city street at night, passing a yellow taxi before continuing toward a distant red traffic light, while the camera tracks behind from a slightly elevated rear angle.\par\medskip
\promptfield{retention\_analysis}<Video~1> (frame-by-frame layout of every class and camera): attribute\_transfer \textendash{} every region keeps its class, outline and place; materials, colours and lighting are not given by <Video~1>; the flat class colours are not shown.\\
<Video~2> ([Shot 1] first frame): fully\_preserved \textendash{} the composition, subjects, materials, colours and lighting of <Video~2> are the target video's first frame.\par\medskip
\promptfield{detailed\_description}The target video is a realistic 3D game render in a dark, foggy night with wet reflective surfaces and soft, diffused street lighting cutting through heavy atmospheric haze. [Shot 1] The shot begins from <Video~2>. The opening frame shows a bright blue compact hatchback with a black roof and rear spoiler positioned in the right lane of a multi-lane urban road, viewed from a slightly elevated rear-three-quarter angle. \ldots
\end{promptcard}
\caption{\textbf{Training caption from CARLA with a semantic map.} The detailed description is truncated.}
\label{fig:cap-carla}
\end{figure}

\subsection{Implementation Details}
\label{app:implementation}

The backbone is the official MiniMax-H3 33B Ref2VA model and stays frozen
throughout. The control LoRA has rank 384 on the attention and
feed-forward layers and is trained with the native flow-matching
objective of the backbone on clips of 158 frames at $1344\times768$ and
24\,fps, using AdamW \citep{loshchilov2019adamw} with a learning rate of $10^{-4}$ for 38,400 steps at a batch of 4.
For structure-guided post-training, the control LoRA is frozen and the
teacher is the backbone with the control LoRA. The four-step LoRA has rank
384, with its first 128 ranks initialized from the pretrained DMAD LoRA
and the rest at zero, and the fake-score model is a separate LoRA
initialized at the teacher. The generator is updated once every five
fake-score updates, with learning rates of $10^{-5}$ for the generator and
$2\times10^{-6}$ for the fake-score model, for 800 iterations at a batch of 4 with $w=1.5$. The conditioning and data ablations use an earlier data mixture.
The causal control LoRA (\cref{sec:causal}) starts from the control LoRA
after 38,400 steps and is trained with teacher forcing for 19,200 steps at a
batch of 2, using AdamW with a learning rate of $3\times10^{-5}$. The
causal acceleration LoRA starts from the bidirectional four-step LoRA and
is trained for 1,600 iterations at a batch of 1 with the causal control LoRA
frozen. The generator and the fake-score model, a rank-128 LoRA initialized
at the teacher, use learning rates of $2\times10^{-5}$ and
$5\times10^{-5}$, with five fake-score updates per generator update.
All schedules are given for one 8-GPU node, as in the released
configurations; the released weights were trained on as many
samples at a larger batch.

\subsection{Evaluation Protocol and Metrics}
\label{app:evaluation}

This section defines the automatic metrics, the automatic score, and the
pairwise judge with its Elo rating. Let $V=\{I_t\}_{t=1}^{T}$ be a rendered
video, $S=\{S_t\}_{t=1}^{T}$ its structural video, and $I^{\mathrm{ref}}$ the
given first frame. $f(\cdot)$ denotes the global DINOv2 ViT-S/14 embedding and $\cos(\cdot,\cdot)$ cosine similarity. Every
metric is computed per case and averaged over the cases of a test set.

\numberedmetriccard{metricquality}{Visual Quality}
Three learned predictors rate the appearance of the rendering alone.

\coloredstep{metricquality}{DOVER.} The overall score of DOVER on the clip, in $[0,1]$.

\coloredstep{metricquality}{Aesthetic Score.} The LAION aesthetic predictor on CLIP ViT-L/14 \citep{radford2021clip} embeddings of two frames per
second, averaged over frames, in $[1,10]$.

\coloredstep{metricquality}{VideoScore2.} The visual-quality rating of
VideoScore2 at 2\,fps with greedy decoding, read
as the expected level $\sum_{k=1}^{5}k\,p(k)$ over its five levels.

\numberedmetriccard{metricstability}{Temporal Stability}
Two measures of frame-to-frame consistency.

\coloredstep{metricstability}{Flicker.} Each frame $I_{t+1}$ is warped onto
$I_t$ with RAFT optical flow \citep{teed2020raft}, giving $\tilde I_{t+1}$, and compared on the
pixels $M_t$ whose forward and backward flows agree, after a Gaussian
low-pass $G$ ($\sigma=2$\,px) so that fine texture is not counted:
\begin{equation}
 E_{\mathrm{flicker}}=\frac{1}{T-1}\sum_{t=1}^{T-1}\frac{1}{|M_t|}
 \sum_{p\in M_t}\bigl|(G*I_t)(p)-(G*\tilde I_{t+1})(p)\bigr|,
\end{equation}
with pixel values in $[0,255]$. Lower is better.

\coloredstep{metricstability}{Adjacent-Frame Similarity.}
$S_{\mathrm{DINO}}=\frac{1}{T-1}\sum_{t=1}^{T-1}\cos\bigl(f(I_t),f(I_{t+1})\bigr)$.

\numberedmetriccard{metricstructure}{Structural Adherence}
Two measures compare the rendering with its structural video.

\begin{figure}[t]
\centering
\begin{promptcard}
\promptfield{subject\_definitions}<Video~1> is a grey shaded rendering of the engine's depth within about 30 metres of the camera, from the same camera as the target video; everything farther, and the sky, is black. It gives the exact shapes, positions and motion of the nearby scene, characters and vehicles, and the camera of every frame.\\
<Video~2> is the first frame of [Shot 1], held still.\par\medskip
\promptfield{summary}[keyframe completion + reference generation] The target video begins from <Video~2> and follows the frame-by-frame structure and motion of <Video~1>. A man in a blue shirt and beige shorts chases a taxi and a black car down a dark city street at night, running along the double yellow lines as the vehicles speed away from him and the camera follows from behind.\par\medskip
\promptfield{retention\_analysis}<Video~1> (frame-by-frame near geometry, motion and camera): attribute\_transfer \textendash{} the shapes, positions and motion of everything within the near range and the camera are followed; the far scene, materials, colours and lighting are not given by <Video~1> and are the target video's own; the grey shading and the black background of <Video~1> are not shown.\\
<Video~2> ([Shot 1] first frame): fully\_preserved \textendash{} the composition, subjects, materials, colours and lighting of <Video~2> are the target video's first frame.\par\medskip
\promptfield{detailed\_description}The target video is a realistic 3D CG game render in nighttime urban lighting with warm sodium-vapor streetlamps and cool distant building illumination. [Shot 1] The shot begins from <Video~2>. The opening frame shows a man in a blue long-sleeve shirt and beige knee-length shorts standing in the middle of a two-lane road marked with double yellow lines, facing away from the camera. \ldots
\end{promptcard}
\caption{\textbf{Training caption from an engine recording with near-range depth.} The detailed description is truncated.}
\label{fig:cap-game}
\end{figure}

\coloredstep{metricstructure}{Motion Error.} CoTracker3 tracks the same $20\times12$ grid of points in
$S$ and in $V$, giving positions $q_j(t)$ and $\hat q_j(t)$. Over the set
$\mathcal R$ of points and frames where the structural track is reliable,
\begin{equation}
 E_{\mathrm{motion}}=\frac{1}{|\mathcal R|\,D}\sum_{(j,t)\in\mathcal R}
 \bigl\|\hat q_j(t)-q_j(t)\bigr\|_2,
\end{equation}
where $D$ is the image diagonal. Lower is better.

\coloredstep{metricstructure}{Edge F1.} On every fourth frame, Canny edges \citep{canny1986computational}
$\mathcal E_t$ of the rendering are matched to the edges $\mathcal E^S_t$ of
the structural video with a tolerance of 2\,px. With precision $P_t$ the
fraction of $\mathcal E_t$ within tolerance of $\mathcal E^S_t$ and recall
$R_t$ the converse, the score averages $2P_tR_t/(P_t+R_t)$ over frames.

\numberedmetriccard{metricfirst}{First-Frame Preservation}
The appearance of the given first frame should persist through the clip.
Over the frames $\mathcal L$ of the last second,
\begin{equation}
 S_{\mathrm{first}}=\frac{1}{|\mathcal L|}\sum_{t\in\mathcal L}
 \cos\bigl(f(I_t),f(I^{\mathrm{ref}})\bigr).
\end{equation}
\numberedmetriccard{metricjudge}{Automatic Score}
Each metric $x$ is mapped to a fixed 0--100 scale,
\begin{equation}
 n(x)=100\,\frac{x-a}{b-a}\ \text{(higher is better)},\qquad
 n(x)=100\,\frac{b-x}{b-a}\ \text{(lower is better)},
\end{equation}
with $[a,b]=[0,1]$ for DOVER, edge F1, adjacent-frame similarity, and
first-frame similarity, $[1,10]$ for the aesthetic score, $[1,5]$ for
VideoScore2, $[0,10]$ for flicker, and $[0,0.5]$ for motion error. The
score of a group is the mean of $n(x)$ over its metrics. For the official
backbone and Wan2.2-VACE-Fun, the structure metrics use the frames they
render.

\begin{figure}[t]
\centering
\begin{promptcard}
\promptfield{subject\_definitions}<Video~1> is an untextured grey block-out of the scene, rendered from the same camera as the target video, giving the layout, the large geometry, the characters' bodies, their motion and the camera of every frame.\\
<Video~2> is the first frame of [Shot 1], held still.\par\medskip
\promptfield{summary}[keyframe completion + reference generation] The target video begins from <Video~2> and follows the frame-by-frame structure and motion of <Video~1>. A man in a dark leather jacket runs through a rubble-strewn, war-torn urban courtyard, vaulting over a low concrete barrier and continuing across debris toward a burning car, while the camera tracks behind and arcs around him.\par\medskip
\promptfield{retention\_analysis}<Video~1> (frame-by-frame layout, geometry, motion and camera): attribute\_transfer \textendash{} the layout, large shapes, character bodies, motion and camera are followed exactly; small objects, materials, colours and lighting are not given by <Video~1> and are the target video's own; the grey look of <Video~1> is not shown.\\
<Video~2> ([Shot 1] first frame): fully\_preserved \textendash{} the composition, subjects, materials, colours and lighting of <Video~2> are the target video's first frame.\par\medskip
\promptfield{detailed\_description}The target video is a realistic 3D game render in bright midday sun with sharp, crisp shadows and warm, slightly desaturated color grading. [Shot 1] The shot begins from <Video~2>. The opening frame shows a man standing with his back to camera in a narrow passage of a destroyed urban environment. \ldots
\end{promptcard}
\caption{\textbf{Training caption from Magpie with a scene block-out.} The detailed description is truncated.}
\label{fig:cap-magpie}
\end{figure}

\numberedmetriccard{metricjudge}{Pairwise Judge and Elo}
Gemini~3.8 Flash receives the structural video, the first
frame, and two renderings, and answers four questions: which rendering has
higher visual quality, which follows the structure and motion more closely,
which keeps the first frame better, and which it would rather ship. Each
pair is judged in both orders, and a case is a win for a method when both
orders prefer it or one prefers it and the other is a tie. With $W$ wins,
$L$ losses, and $N$ cases, the net score is $(W-L)/N\in[-1,1]$.

\coloredstep{metricjudge}{Elo Rating.} From the verdicts between all pairs
of methods on a test set we fit one Bradley--Terry model per judge question,
\begin{equation}
 \Pr(i\succ j)=\frac{1}{1+10^{(R_j-R_i)/400}},
\end{equation}
by maximum likelihood, counting a tie as half a win for each side and
adding one virtual tie per method against a fixed reference, so that a
method that loses every comparison still has a finite rating. Ratings are
shifted so that the mean of the compared methods is 1000.

\coloredstep{metricjudge}{Judge Inputs.} The three videos are resized to
$672\times384$ at 12\,fps. Each input is preceded by a label (reference structure, first
frame, video A, video B). The prompt asks the judge to assess visual
quality on the two videos alone, to ignore the crude appearance of the
structure, and not to favor either position, and requests a JSON answer
with a reason of at most 40 words.

\coloredstep{metricjudge}{Human Study.}
\label{sec:human-study}
In a blind study, human raters compared pairs of renderings shown side by
side in random order without method names, and chose one of the two videos
or declared a tie. Their preferences agree with the judge in direction.

\subsection{Compared Methods and Generation Settings}
\label{app:baselines}

We group the compared methods into structure-conditioned renderers,
baselines, and existing acceleration LoRAs. Each method runs with its
released resolution, frame count, sampler, and number of steps, and its
outputs are resized to $1344\times768$ at 24\,fps before scoring.

\subsubsection{Structure-Conditioned Renderers}

\coloredstep{methodrender}{LynnReal-Omni}
is a four-step omni-reference model on the same backbone. We render at
$1344\times768$ and 24\,fps with 4 steps, and the first frame enters as a
style reference.

\begin{figure}[t]
\centering
\begin{promptcard}
\promptfield{subject\_definitions}<Video~1> is the objects of the scene as flat-coloured masks, one colour per object, with the skeletons of the characters drawn on top, frame by frame, from the same camera as the target video.\\
<Video~2> is the first frame of [Shot 1], held still.\par\medskip
\promptfield{summary}[keyframe completion + reference generation] The target video begins from <Video~2> and follows the frame-by-frame structure and motion of <Video~1>. The camera glides forward at low height through a sun-bleached desert corridor between ancient stone ruins, revealing weathered block walls, sandy ground, and distant architectural fragments under harsh midday light.\par\medskip
\promptfield{retention\_analysis}<Video~1> (frame-by-frame object outlines, character poses and camera): attribute\_transfer \textendash{} every object keeps its outline and place, every character keeps its pose and motion; materials, colours and lighting are not given by <Video~1>; the flat mask colours and skeleton lines are not shown.\\
<Video~2> ([Shot 1] first frame): fully\_preserved \textendash{} the composition, subjects, materials, colours and lighting of <Video~2> are the target video's first frame.\par\medskip
\promptfield{detailed\_description}The target video is a stylized 3D game render with warm, slightly desaturated tones and hard directional sunlight creating deep shadows. [Shot 1] The shot begins from <Video~2>. The opening frame shows a low-angle view across rippled orange-tan sand, scattered with small dark pebbles, dried mud clumps, faint tire-like grooves, and isolated patches where darker wet sand shows through. \ldots
\end{promptcard}
\caption{\textbf{Training caption from UnrealZoo with object masks and skeletons.} The detailed description is truncated.}
\label{fig:cap-unrealzoo}
\end{figure}

\coloredstep{methodrender}{StreamRender-H3}
is a same-backbone streaming renderer trained on semantic maps of a racing
game. We use its released settings with an $832\times480$ reference and two-step streaming.

\coloredstep{methodrender}{Streaming CWM} is a
same-backbone renderer for the proxy videos of a code world model, trained
on CARLA. We use its released four-step LoRA, resize the structure to its
$336\times192$ proxy input, and concatenate ten 17-frame chunks and trim
them to 158 frames.

\coloredstep{methodrender}{PAI ControlNet-Union}
is a multi-condition ControlNet for video. We render 158 frames at
$1280\times704$ with 40 steps and guidance 1.0, with the first frame as its inpainting input.

\coloredstep{methodrender}{Cosmos-Transfer2.5}
transfers control videos into realistic video. We render 93-frame chunks at
$1280\times704$ with 35 steps and guidance 3, and the first frame serves as
a style reference.

\coloredstep{methodrender}{Wan2.2-VACE-Fun}
is an all-in-one video creation and editing model. We render 81 frames at
$832\times480$ and 16\,fps (the first five seconds) with 50 steps and
guidance 5.0, with the first frame as the start image.

\begin{figure}[t]
\centering
\begin{promptcard}
\promptfield{subject\_definitions}<Video~1> is a grey 3D body mesh of the main character on a plain white background, rendered from the same camera as the target video; it gives the character's position, size, body pose and motion in every frame, and nothing of the surroundings.\\
<Video~2> is the first frame of [Shot 1], held still.\par\medskip
\promptfield{summary}[keyframe completion + reference generation] The target video begins from <Video~2> and follows the frame-by-frame structure and motion of <Video~1>. A man in western attire walks down a grassy slope toward a misty river in a forested valley. The camera follows him from behind at a consistent distance, maintaining a third-person perspective as he navigates the terrain.\par\medskip
\promptfield{retention\_analysis}<Video~1> (frame-by-frame body pose and motion of the main character): attribute\_transfer \textendash{} the main character's position, size, pose and motion are followed; the character's clothing, face and hair, the surroundings, the camera's view of the scene, materials, colours and lighting are not given by <Video~1> and are the target video's own; the grey mesh and the white background of <Video~1> are not shown.\\
<Video~2> ([Shot 1] first frame): fully\_preserved \textendash{} the composition, subjects, materials, colours and lighting of <Video~2> are the target video's first frame.\par\medskip
\promptfield{detailed\_description}The target video is a realistic 3D rendered game scene in overcast, misty daylight with soft, diffused lighting and no hard shadows. [Shot 1] The shot begins from <Video~2>. The composition centers on a man viewed from behind, walking down a grassy hillside toward a wide, fast-flowing river. \ldots
\end{promptcard}
\caption{\textbf{Training caption from a YouTube video with an SMPL body.} The detailed description is truncated.}
\label{fig:cap-smpl}
\end{figure}

\subsubsection{Baselines}

\coloredstep{methodbase}{Official backbone} is the
MiniMax-H3 Ref2VA model without our adapters, which receives the
structural video as an ordinary reference video. We render 124 frames at
$1344\times768$ and 24\,fps with 50 steps.

\coloredstep{methodbase}{Ours (Stage~1)} is our control model before
post-training, sampled with 50 steps.

\subsubsection{Existing Acceleration LoRAs}

\coloredstep{methodaccel}{LongLive-Plug, DMAD, and FastH3} are four-step acceleration LoRAs trained on the backbone. Each is attached
to our Stage~1 control LoRA without further training and sampled with
four steps.

\begin{figure}[t]
\centering
\begin{promptcard}
\promptfield{subject\_definitions}<Video~1> is the first frame of [Shot 1], held still.\par\medskip
\promptfield{summary}[keyframe completion] The target video begins from <Video~1>. An armored hunter wielding a massive bone greatsword battles a serpentine monster in a dark rocky cavern. The hunter charges, dodges, and unleashes a fiery charged attack against the coiling creature, with the camera tracking dynamically around the combat.\par\medskip
\promptfield{retention\_analysis}<Video~1> ([Shot 1] first frame): fully\_preserved \textendash{} the composition, subjects, materials, colours and lighting of <Video~1> are the target video's first frame.\par\medskip
\promptfield{detailed\_description}The target video is a realistic 3D CG game render in dim, cool-toned cavern light with soft volumetric fog and muted shadows. [Shot 1] The shot begins from <Video~1>. The scene opens on a hunter in brown and bronze segmented plate armor standing on dark wet stone, facing a massive serpentine monster with a broad armored head, glowing yellow eyes, and multiple long black tentacles lined with yellow-gold ridges. \ldots
\end{promptcard}
\caption{\textbf{Training caption from WildWorld with a first frame only.} The detailed description is truncated.}
\label{fig:cap-wildworld}
\end{figure}

\begin{table*}[t]
\begin{minipage}[t]{0.485\linewidth}
\centering
\begin{minipage}[t][50pt][t]{\linewidth}
\caption{\textbf{Comparison on unseen representations.} Gemini Elo ratings and automatic totals on 74 cases. The causal variant is rated only against ours. Bold and underlined italics mark the best and second-best values.}
\label{tab:unseen}
\end{minipage}\par
\tablefont
\renewcommand{\resulthead}[1]{{\fontsize{6.5}{8}\selectfont\sffamily\bfseries\shortstack{#1}}}
\setlength{\tabcolsep}{1.2pt}
\renewcommand{\arraystretch}{1.2}
\begin{tabularx}{\linewidth}{l c*{5}{>{\centering\arraybackslash}X}}
\toprule
\rowcolor{tabhead}
\resulthead{Method} & \resulthead{Steps} & \resulthead{Over.} & \resulthead{Qual.} & \resulthead{Struct.} & \resulthead{First} & \resulthead{Auto.}\\
\midrule
\rowcolor{tabgroup}
\multicolumn{7}{l}{\textbf{Baselines}}\\
Official backbone & $50$ & $998$ & $1038$ & $722$ & $1120$ & $69.9$\\
\rowcolor{tabstripe}
\textbf{Ours (Stage~1)} & $50$ & $1318$ & $1272$ & \underline{$\mathit{1280}$} & $1325$ & \underline{$\mathit{70.4}$}\\
\midrule
\rowcolor{tabgroup}
\multicolumn{7}{l}{\textbf{Structure-conditioned renderers}}\\
Wan2.2-VACE-Fun & $50$ & $519$ & $622$ & $598$ & $508$ & $65.8$\\
\rowcolor{tabstripe}
PAI ControlNet-Union & $40$ & $1215$ & $1224$ & $1046$ & $1103$ & $68.7$\\
Cosmos-Transfer2.5 & $35$ & $519$ & $500$ & $1105$ & $508$ & $64.0$\\
\rowcolor{tabstripe}
LynnReal-Omni & $4$ & $\mathbf{1409}$ & \underline{$\mathit{1369}$} & $1275$ & $\mathbf{1405}$ & $69.7$\\
Streaming CWM & $4$ & $1098$ & $1095$ & $891$ & $1135$ & $70.1$\\
\rowcolor{tabstripe}
StreamRender-H3 & $2$ & $519$ & $500$ & $722$ & $508$ & $62.3$\\
\midrule
\textbf{Ours (causal)} & $4$ & $1132$ & $1068$ & $1201$ & $1103$ & $69.8$\\
\rowcolor{tabours}
\textbf{Ours} & $4$ & \underline{$\mathit{1403}$} & $\mathbf{1381}$ & $\mathbf{1360}$ & \underline{$\mathit{1390}$} & $\mathbf{70.6}$\\
\bottomrule
\end{tabularx}
\end{minipage}
\hfill
\begin{minipage}[t]{0.485\linewidth}
\centering
\begin{minipage}[t][50pt][t]{\linewidth}
\caption{\textbf{Automatic scores of the conditioning and data ablations.} Group scores and their mean (0--100) for \cref{tab:injection,tab:abl-data}. Higher is better; bold marks the best value within each block.}
\label{tab:injection-full}
\end{minipage}\par
\tablefont
\renewcommand{\resulthead}[1]{{\fontsize{6.5}{8}\selectfont\sffamily\bfseries\shortstack{#1}}}
\setlength{\tabcolsep}{1.2pt}
\renewcommand{\arraystretch}{1.347}
\begin{tabularx}{\linewidth}{l *{5}{>{\centering\arraybackslash}X}}
\toprule
\rowcolor{tabhead}
\resulthead{Model} & \resulthead{Qual.} & \resulthead{Stab.} & \resulthead{Ctrl.} & \resulthead{First} & \resulthead{Auto.}\\
\midrule
\rowcolor{tabgroup}
\multicolumn{6}{l}{\textbf{Conditioning design}}\\
ControlNet branch & $\mathbf{60.2}$ & $90.5$ & $46.8$ & $66.4$ & $66.0$\\
Half resolution & $59.5$ & $91.5$ & $46.6$ & $\mathbf{67.0}$ & $66.2$\\
Separate RoPE & $59.1$ & $91.5$ & $46.4$ & $\mathbf{67.0}$ & $66.0$\\
\rowcolor{tabours}
\textbf{Ours (full res.)} & $59.6$ & $\mathbf{91.6}$ & $\mathbf{48.0}$ & $66.8$ & $\mathbf{66.5}$\\
\midrule
\rowcolor{tabgroup}
\multicolumn{6}{l}{\textbf{Training data (half resolution)}}\\
All data & $59.5$ & $91.5$ & $46.6$ & $67.0$ & $66.2$\\
Structured only & $59.7$ & $91.2$ & $47.3$ & $66.8$ & $66.2$\\
$1/10$ data & $59.6$ & $91.4$ & $46.5$ & $\mathbf{67.1}$ & $66.1$\\
New seed & $58.9$ & $91.5$ & $46.4$ & $66.9$ & $65.9$\\
\rowcolor{tabours}
\textbf{W/o misaligned} & $\mathbf{59.9}$ & $\mathbf{91.6}$ & $\mathbf{47.4}$ & $66.9$ & $\mathbf{66.5}$\\
\bottomrule
\end{tabularx}
\end{minipage}

\end{table*}

\section{Data and Prompt Specifications}
\label{app:data-prompts}

\subsection{Training Captions}

All structural references enter the same reference pathway, and the
caption states what each of them constrains. Every caption has a fixed
definition and retention line for each supplied reference, which tells the
model which information the reference gives and that its look is not
shown, followed by a summary and a detailed description of the target
video. \Cref{fig:cap-carla,fig:cap-game,fig:cap-magpie,fig:cap-unrealzoo,fig:cap-smpl,fig:cap-wildworld} show one real caption from six sources each.

\begin{figure*}[t]
\centering
\includegraphics[width=\linewidth]{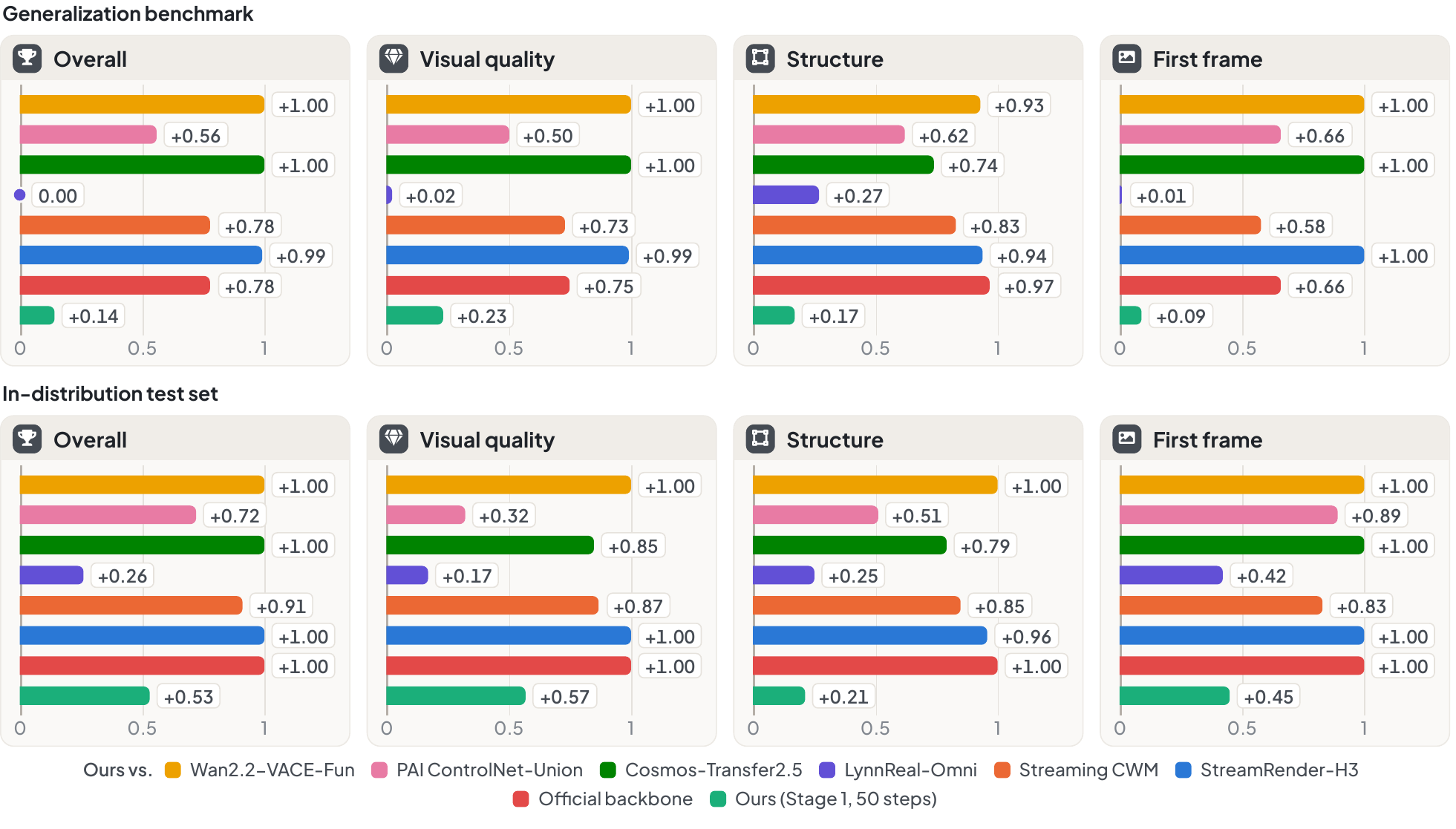}
\caption{\textbf{Pairwise margin of ours over each compared method.} Net
score of ours against each indicated method for each judge question on
the two test sets. Positive values favor ours.}
\label{fig:netscore}
\end{figure*}

\subsection{Caption Validation}

The captioning pipeline uses Kimi K2.6. For every caption, validators require
all six fields, only defined reference tags, a 350--500-word body, no Chinese
outside quotation marks, and increasing shot times. When a first frame is
provided, the first shot must begin from that reference. Captions that fail these checks are regenerated, with at most three retries.

\subsection{Source Audit Details}
\label{app:source-audits}

\noindent\textbf{Source Alignment.}
On Magpie, the synchronization audit retains 32,027 of 44,266 clips. Shot
changes, static content, structural mismatch, and darkness account for most
rejections, followed by near-duplicates, black frames, and unmatched
jumps. For the engine recordings, depth, masks, camera poses, and skeletons
are aligned to the RGB timestamps,
80 of 84 motion windows have their best offset at zero, and
depth-and-camera reprojection has a median error of 0.06\%. The near-range
depth conversion retains 13,700 of 13,702 clips. A finer proxy built from
0.25\,m blocks was also considered but not adopted, because its flicker is
four to seven times higher.

\begin{table}[t]
\centering
\caption{\textbf{Automatic scores on both test sets.} Four group scores and their equally weighted total (0--100, higher is better) on the generalization benchmark (208 cases) and in-distribution test set (53 clips). Bold and underlined italics mark the best and second-best values.}
\label{tab:absolute}
\tablefont
\setlength{\tabcolsep}{1pt}
\renewcommand{\arraystretch}{1.2}
\begin{tabularx}{\linewidth}{l c*{10}{>{\centering\arraybackslash}X}}
\toprule
\rowcolor{tabhead}
& & \multicolumn{5}{c}{\textit{Generalization benchmark}} & \multicolumn{5}{c}{\textit{In-distribution test set}}\\
\rowcolor{tabhead}
\resulthead{Method} & \resulthead{Steps} & \resulthead{Quality} & \resulthead{Stability} & \resulthead{Control} & \resulthead{First fr.} & \resulthead{Auto.} & \resulthead{Quality} & \resulthead{Stability} & \resulthead{Control} & \resulthead{First fr.} & \resulthead{Auto.}\\
\midrule
\rowcolor{tabgroup}
\multicolumn{12}{l}{\textbf{Baselines}}\\
Official backbone & $50$ & $63.8$ & $90.7$ & $55.6$ & \underline{$\mathit{69.3}$} & $69.8$ &  $54.4$ & \underline{$\mathit{91.4}$} & $44.4$ & \underline{$\mathit{69.1}$} & $64.8$\\
\rowcolor{tabstripe}
\textbf{Ours (Stage~1)} & $50$ & \underline{$\mathit{69.3}$} & $90.7$ & $57.8$ & $68.8$ & $\mathbf{71.6}$ &  $60.3$ & \underline{$\mathit{91.4}$} & $43.8$ & $\mathbf{69.4}$ & $66.2$\\
\midrule
\rowcolor{tabgroup}
\multicolumn{12}{l}{\textbf{Structure-conditioned renderers}}\\
Wan2.2-VACE-Fun & $50$ & $60.3$ & $87.6$ & \underline{$\mathit{62.6}$} & $50.3$ & $65.2$ &  $55.9$ & $85.4$ & $\mathbf{59.9}$ & $50.8$ & $63.0$\\
PAI ControlNet-Union & $40$ & $67.5$ & $91.4$ & $56.0$ & $62.9$ & $69.4$ &  $61.5$ & $90.8$ & $44.4$ & $61.7$ & $64.6$\\
Cosmos-Transfer2.5 & $35$ & $52.3$ & $90.0$ & $\mathbf{69.5}$ & $47.2$ & $64.7$ &  $57.5$ & $89.6$ & \underline{$\mathit{58.3}$} & $53.3$ & $64.7$\\
LynnReal-Omni & $4$ & $\mathbf{70.7}$ & $86.4$ & $55.4$ & $67.5$ & $70.0$ &  $\mathbf{64.8}$ & $86.5$ & $39.1$ & $66.4$ & $64.2$\\
Streaming CWM & $4$ & $67.9$ & $\mathbf{92.3}$ & $50.3$ & $\mathbf{72.1}$ & $70.7$ &  $59.4$ & $\mathbf{92.1}$ & $36.3$ & \underline{$\mathit{69.1}$} & $64.2$\\
StreamRender-H3 & $2$ & $61.9$ & $83.4$ & $54.8$ & $55.4$ & $63.9$ &  $55.3$ & $82.3$ & $43.7$ & $54.8$ & $59.0$\\
\textbf{Ours (causal)} & $4$ & $68.8$ & \underline{$\mathit{91.6}$} & $60.0$ & $64.3$ & \underline{$\mathit{71.2}$} &  $60.3$ & \underline{$\mathit{91.4}$} & $48.5$ & $66.3$ & \underline{$\mathit{66.6}$}\\
\rowcolor{tabours}
\textbf{Ours} & $4$ & $69.1$ & $89.8$ & $61.0$ & $66.7$ & $\mathbf{71.6}$ &  \underline{$\mathit{61.9}$} & $89.8$ & $48.9$ & $67.5$ & $\mathbf{67.0}$\\
\bottomrule
\end{tabularx}
\end{table}

\noindent\textbf{YouTube Videos.}
We keep 137 of 559 videos with a clear protagonist. Our target stream and the stream used for SMPL fitting differ
by about 2 grey levels per pixel on a 0--255 scale, and the mesh videos
match the target in crop, resolution, length, and frame rate.

\begin{figure}[t]
\centering
\includegraphics[width=\linewidth]{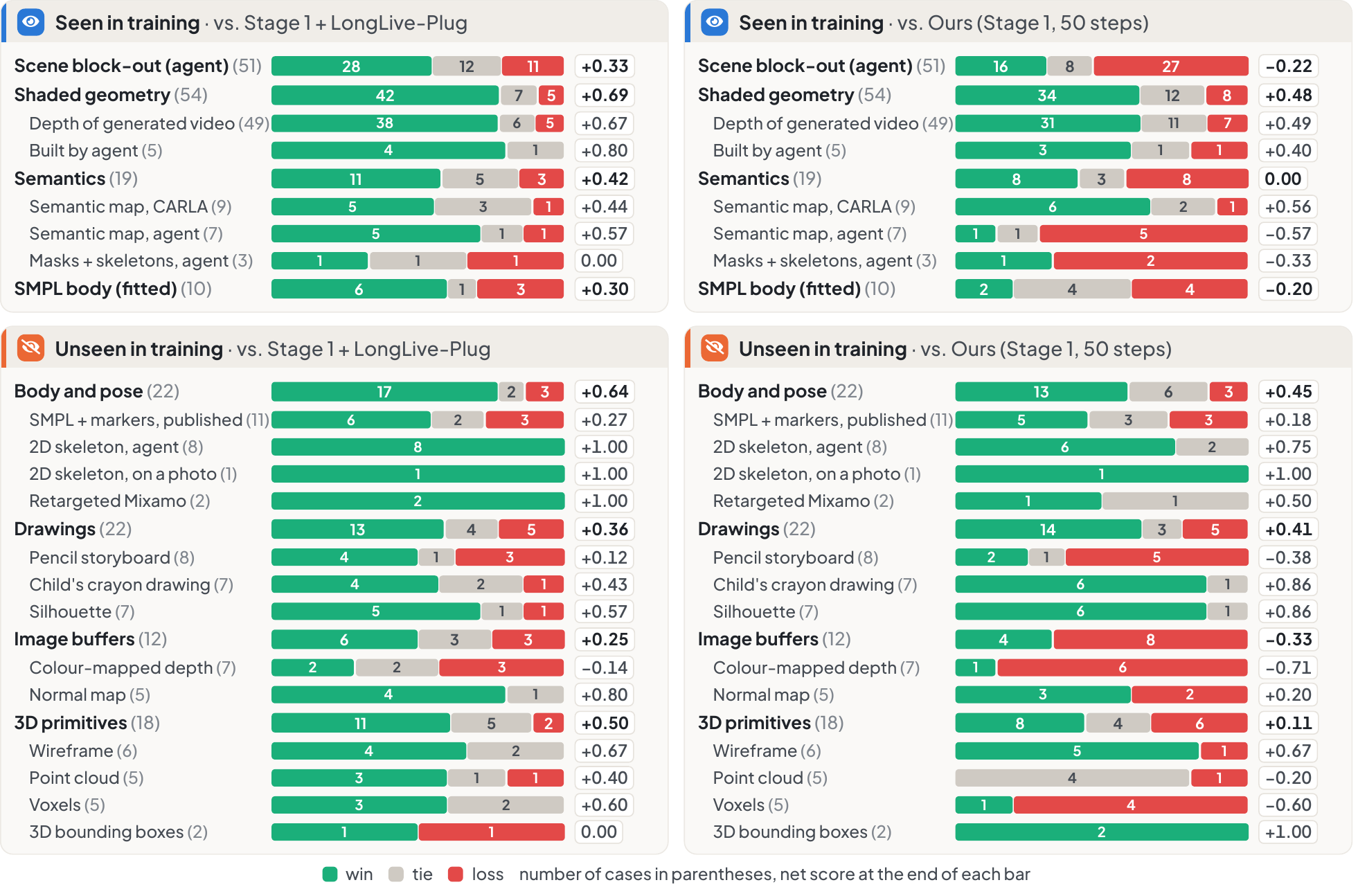}
\caption{\textbf{Results by representation and source on the generalization
benchmark.} Each bar splits the cases of one representation into wins,
ties, and losses of ours against Stage~1 + LongLive-Plug (left) and against
our 50-step Stage~1 model (right), with the number of cases in parentheses
and the net score at the end of the bar. Bold rows match the categories in
\cref{fig:by-repr}, with individual representations and sources below. Positive values favor ours.}
\label{fig:per-condition}
\end{figure}

\section{Additional Experimental Results}
\label{app:more-results}

\subsection{Detailed Results}
\label{app:results}

We extend the main comparison with ratings on unseen representations,
direct pairwise margins, and automatic scores grouped by evaluation dimension. We then
break down representation groups to compare four-step rendering with
the full-step control model.

\noindent\textbf{Generalization to Unseen Representations.}
\Cref{tab:unseen} restricts the comparison to the 74 cases whose structural
representation never occurs in training. \project achieves the
\textbf{\textit{highest quality and structure ratings}}, with a structure
rating of 1360 against 1275 for LynnReal-Omni. Its overall rating is only
six points below that competitor, while the quality rating is higher and
the first-frame rating is slightly lower. The clearest advantage is thus
following unfamiliar structure while producing competitive appearance.
The four-step model also improves on Stage~1 on all four questions,
showing that the benefit of structure-guided post-training extends to
representations absent from the training corpus. It also leads the total automatic
score at 70.6, followed by Stage~1 at 70.4.

\noindent\textbf{Pairwise Comparison with Existing Methods.}
\Cref{fig:netscore} reports the normalized win--loss margin of ours against each
method, complementing the ratings in \cref{tab:main}. Every margin is
nonnegative on both test sets. Against LynnReal-Omni, the overall
comparison is even on the generalization benchmark, but the structure
margin reaches $+0.27$, and all four questions favor ours on the
in-distribution test set. Against Stage~1, quality improves alongside
structure and first-frame preservation on both sets. These direct
comparisons support the same conclusion as the Elo ratings:
\textbf{\textit{four-step rendering improves the controlled output}},
with the strongest gain over the closest external competitor in structure.

\noindent\textbf{Results by Representation and Source.}
\Cref{fig:per-condition} expands the categories in \cref{fig:by-repr}
into individual representations and sources. Compared with
Stage~1 + LongLive-Plug, ours has a positive margin in every broad group,
with particularly strong results on pose inputs and shaded geometry.
Against our 50-step Stage~1 model, the four-step model wins on six of eight
agent-built skeleton cases and on six of seven silhouettes, but loses
on colour-mapped depth, voxels, and pencil storyboards. Thus, the gains
on unseen inputs cover several distinct formats, while some individual
formats still favor full-step sampling. The category totals and the
case counts make both patterns visible beyond the aggregate comparison.

\begin{table*}[t]
\begin{minipage}[t]{0.485\linewidth}
\centering
\begin{minipage}[t][50pt][t]{\linewidth}
\caption{\textbf{Automatic scores of the structure-guidance ablation.} Group scores and their mean (0--100, higher is better) for \cref{tab:abl-guidance}. Bold marks the best per block.}
\label{tab:abl-guidance-full}
\end{minipage}\par
\tablefont
\renewcommand{\resulthead}[1]{{\fontsize{6.5}{8}\selectfont\sffamily\bfseries\shortstack{#1}}}
\setlength{\tabcolsep}{1.2pt}
\renewcommand{\arraystretch}{1.2}
\begin{tabularx}{\linewidth}{l *{5}{>{\centering\arraybackslash}X}}
\toprule
\rowcolor{tabhead}
\resulthead{Model} & \resulthead{Qual.} & \resulthead{Stab.} & \resulthead{Ctrl.} & \resulthead{First} & \resulthead{Auto.}\\
\midrule
\rowcolor{tabgroup}
\multicolumn{6}{l}{\textbf{Self-distillation}}\\
$w=1$ & $68.6$ & $\mathbf{91.1}$ & $59.4$ & $\mathbf{68.7}$ & $71.9$\\
$w=2$ & $\mathbf{69.3}$ & $90.5$ & $\mathbf{61.2}$ & $67.6$ & $\mathbf{72.1}$\\
\midrule
\rowcolor{tabgroup}
\multicolumn{6}{l}{\textbf{LongLive-Plug start}}\\
$w=1$ & $68.0$ & $\mathbf{91.4}$ & $59.1$ & $\mathbf{68.2}$ & $71.7$\\
$w=1.5$ & $\mathbf{69.3}$ & $90.8$ & $60.2$ & $68.0$ & $\mathbf{72.1}$\\
$w=2$ & $69.2$ & $90.7$ & $\mathbf{60.4}$ & $68.1$ & $\mathbf{72.1}$\\
\midrule
\rowcolor{tabgroup}
\multicolumn{6}{l}{\textbf{DMAD start}}\\
$w=1$ & $68.0$ & $\mathbf{91.8}$ & $59.2$ & $\mathbf{68.1}$ & $\mathbf{71.8}$\\
\rowcolor{tabours}
\textbf{Ours ($\boldsymbol{w=1.5}$)} & $\mathbf{69.1}$ & $89.8$ & $\mathbf{61.0}$ & $66.7$ & $71.6$\\
\bottomrule
\end{tabularx}
\end{minipage}
\hfill
\begin{minipage}[t]{0.485\linewidth}
\centering
\begin{minipage}[t][50pt][t]{\linewidth}
\caption{\textbf{Automatic scores of the post-training ablation.} Group scores and their mean (0--100) for \cref{tab:abl-design-main}. Ours uses DMAD, $w=1.5$, and four steps. Higher is better, with the best in bold.}
\label{tab:abl-design}
\end{minipage}\par
\tablefont
\renewcommand{\resulthead}[1]{{\fontsize{6.5}{8}\selectfont\sffamily\bfseries\shortstack{#1}}}
\setlength{\tabcolsep}{1.2pt}
\renewcommand{\arraystretch}{1.2}
\begin{tabularx}{\linewidth}{l *{5}{>{\centering\arraybackslash}X}}
\toprule
\rowcolor{tabhead}
\resulthead{Model} & \resulthead{Qual.} & \resulthead{Stab.} & \resulthead{Ctrl.} & \resulthead{First} & \resulthead{Auto.}\\
\midrule
\rowcolor{tabgroup}
\multicolumn{6}{l}{\textbf{Starting point, no post-training}}\\
LongLive-Plug & $69.1$ & $91.1$ & $59.7$ & $\mathbf{68.7}$ & $\mathbf{72.1}$\\
FastH3 & $69.4$ & $90.5$ & $59.8$ & $67.8$ & $71.9$\\
DMAD & $68.8$ & $\mathbf{91.3}$ & $60.4$ & $68.1$ & $\mathbf{72.1}$\\
\midrule
\rowcolor{tabgroup}
\multicolumn{6}{l}{\textbf{Guidance strength}}\\
$w=2$ & $69.3$ & $88.8$ & $60.9$ & $66.3$ & $71.3$\\
\midrule
\rowcolor{tabgroup}
\multicolumn{6}{l}{\textbf{Sampling steps}}\\
2 steps & $69.1$ & $89.9$ & $\mathbf{61.3}$ & $65.7$ & $71.5$\\
8 steps & $\mathbf{69.5}$ & $89.7$ & $61.2$ & $66.0$ & $71.6$\\
\rowcolor{tabours}
\textbf{Ours} & $69.1$ & $89.8$ & $61.0$ & $66.7$ & $71.6$\\
\bottomrule
\end{tabularx}
\end{minipage}

\end{table*}

\noindent\textbf{Comparison on Automatic Metrics.}
\Cref{tab:absolute} compares quality, stability, control, and first-frame
preservation on both test sets. Our method has the highest total on the
in-distribution test set and matches Stage~1 at the reported precision on
the generalization benchmark. Individual renderers lead different groups:
LynnReal-Omni leads quality, Cosmos-Transfer2.5 and Wan2.2-VACE-Fun lead
control on the two sets, and Streaming CWM leads stability but has the
lowest control scores. Our model balances quality and structural agreement.
Compared with Stage~1, post-training improves control on both sets and
quality on the in-distribution test set, while stability and first-frame
preservation decrease. These changes leave the generalization total nearly
unchanged and raise the in-distribution total.

\begin{figure}[t]
\centering
\includegraphics[width=\linewidth]{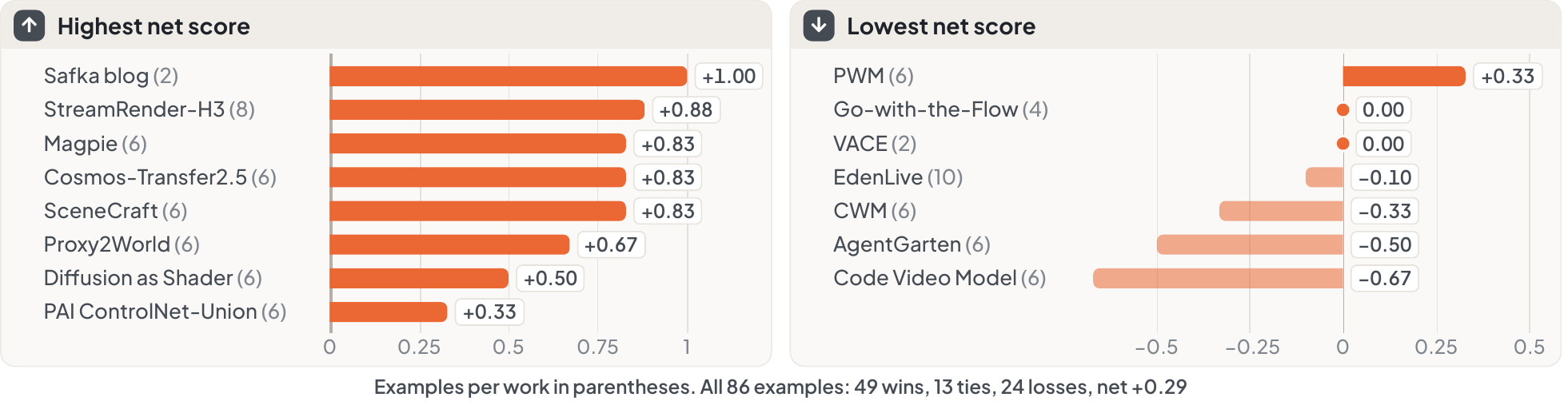}
\caption{\textbf{Visual comparison with other methods.} Overall net score of
ours against the outputs of 15 works on demonstration cases from their
project pages. Positive values favor ours.}
\label{fig:demos}
\end{figure}

\noindent\textbf{Visual Comparison with Other Methods.}
\Cref{fig:demos} summarizes comparisons on 86 demonstration cases from
15 works. We render each case using the structural input shared on its
project page and the provided first frame, or the first frame of its
output video when no separate reference is available. We compare our
outputs with the original results, presenting each pair to the judge in
both orders. The comparison covers StreamRender-H3, Magpie,
Cosmos-Transfer2.5, SceneCraft, Proxy2World, Diffusion as Shader,
PAI ControlNet-Union, Programmable World Model, the neural rendering
examples of \citet{safka2026neuralrendering}, Go-with-the-Flow, VACE,
EdenLive, CWM, AgentGarten, and Code Video Model. Across these cases,
ours wins 49 comparisons, ties 13, and loses 24, giving a net score of
$+0.29$. The positive margin shows that our model also competes well on
cases selected by other methods.

\subsection{Additional Ablations}
\label{app:ablations}

We report the four automatic group scores and their total for the main-text ablations,
focusing on how each design affects appearance, temporal consistency,
and agreement with the structural input.

\noindent\textbf{Effects of Structure Guidance.}
\Cref{tab:abl-guidance-full} supplements \cref{tab:abl-guidance} with
automatic scores. Guidance improves quality and control from every
starting point, consistent with a teacher that more strongly follows the
structural input. From the DMAD start, control rises from 59.2 to 61.0
and quality from 68.0 to 69.1, while stability and first-frame preservation
fall. These changes slightly lower the equally weighted total, even as
the pairwise ratings improve. The automatic groups expose this
tradeoff, while the higher Elo ratings reflect the judge's joint
preference over appearance and structure.

\noindent\textbf{Post-Training and Sampling Settings.}
\Cref{tab:abl-design} gives group scores for the settings in
\cref{tab:abl-design-main}. Post-training the DMAD composition raises
quality and control, while its lower stability and first-frame score
reduce the automatic total. Increasing guidance to $w=2$ further lowers
stability with little change in control. Eight sampling steps improve
quality slightly over four, but reduce first-frame preservation and
leave the total unchanged at the reported precision. Two steps have the
highest control score but a lower first-frame score. Four steps achieve
the highest overall, quality, and first-frame Elo ratings among these
settings, balancing rendering quality with sampling cost.

\noindent\textbf{Conditioning Design.}
\Cref{tab:injection-full} collects the group scores of
\cref{tab:injection,tab:abl-data} and adds their automatic totals.
The native full-resolution pathway improves stability and control over
the ControlNet branch, with a small reduction in quality. Increasing
structural resolution raises control from 46.6 to 48.0 and the total from
66.2 to 66.5, with little change in the other groups. At half resolution,
shared RoPE also improves quality and control over separate positions.
The consistent control gains support aligning reference tokens with the
target frames and preserving the spatial detail of the structural input.

\noindent\textbf{Training Data.}
The lower block of \cref{tab:injection-full} compares data variants.
Keeping only structured samples raises control but lowers stability and
first-frame preservation, leaving the automatic total unchanged at the
reported precision. Removing misaligned structural supervision improves
quality, stability, and control over using all data, raising the total
from 66.2 to 66.5. The corpus therefore benefits from combining reliable
structural supervision with appearance-only samples. Reducing the data
to one tenth has a smaller effect than the variation between full-data
runs, consistent with prioritizing alignment over data volume.

\solarreferences{main}
\end{document}